\documentclass{article}        

\usepackage[utf8]{inputenc}
\usepackage{times}
\usepackage{helvet}
\usepackage{courier}
\usepackage{amsmath}
\usepackage{amsfonts}
\usepackage{graphicx}
\usepackage{algorithmicx}
\usepackage[ruled]{algorithm}
\usepackage{algpseudocode}

\newcommand{\cut}[1]{}

\newcommand{\Rkh}[1]{{#1}}

\newcommand{\BRkh}[2]{{#2}}

\newcommand{\CRkh}[1]{{#1}}

\newcommand{\CP}[1]{}

\newcommand{\CPP}[1]{}

\def\addsymbol #1: #2#3{$#1$ \> \parbox{5in}{#2 \dotfill \pageref{#3}}\\}

\begin{document} 

\title{A New Smooth Approximation to the Zero One Loss with a Probabilistic Interpretation 
}


\author{Md Kamrul Hasan,
        Christopher J. Pal \\
D\'{e}partement de g\'{e}nie informatique et g\'{e}nie logiciel\\
      \'{E}cole Polytechnique Montr\'{e}al\\
      Montr\'{e}al, Qu\'{e}bec, Canada, H3T 1J4\\
md-kamrul.hasan@polymtl.ca, christopher.pal@polymtl.ca       \\
}


\cut{\institute{Md Kamrul Hasan, Christopher Pal \at
              D\'{e}partement de g\'{e}nie informatique et g\'{e}nie logiciel\\
      \'{E}cole Polytechnique Montr\'{e}al\\
      Montr\'{e}al, Qu\'{e}bec, Canada, H3T 1J4\\
\email{md-kamrul.hasan@polymtl.ca}            \\
\email{christopher.pal@polymtl.ca}
}
} 

\date{}

\maketitle

\begin{abstract}
We examine a new form of smooth approximation to the zero one loss in which learning is performed using a reformulation of the widely used logistic function. Our approach is based on using the posterior mean of a novel generalized Beta-Bernoulli formulation. This leads to a generalized logistic function that approximates the zero one loss, but retains a probabilistic formulation conferring a number of useful properties. The approach is easily generalized to kernel logistic regression and easily integrated into methods for structured prediction. We present experiments in which we learn such models using an optimization method consisting of a combination of gradient descent and coordinate descent using localized grid search so as to escape from local minima. Our experiments indicate that optimization quality is improved when learning meta-parameters are themselves optimized using a validation set. Our experiments show improved performance relative to widely used logistic and hinge loss methods on a wide variety of problems ranging from standard UC Irvine and libSVM evaluation datasets to product review predictions and a visual information extraction task. 
We observe that the approach: 1) is more robust to outliers compared to the logistic and hinge losses; 2) outperforms comparable logistic and max margin models on larger scale benchmark problems; 3)
\Rkh{when combined with Gaussian- Laplacian mixture prior on parameters 
the kernelized version of our formulation yields sparser solutions than Support Vector Machine classifiers; and 4) when integrated into a probabilistic structured prediction technique our approach provides more accurate probabilities yielding improved inference and increasing information extraction performance.}
%
%
\end{abstract}

\section{Introduction}

Loss function minimization is a standard way of solving many important learning problems. In the classical statistical literature, this is known as Empirical Risk Minimization (ERM) \cite{vapnik2000nature}, where learning is performed by minimizing the average risk or loss over the training data. Formally, this is represented as 
\begin{equation}
    f^*=\underset{f\in F}{\min} \frac{1}{n}\sum_i^n L(f(\textbf{x}_i),t_i) 
\label{eqn:loss_f}
\end{equation}
where, $f\in F$ is a model, $\textbf{x}_i$ is the i$^th$ input feature vector with label $t_i$, there are $n$ pairs of features and labels, 
and $L(f(\textbf{x}_i),t_i)$ is the loss for the model output $f(\textbf{x}_i)$. 
Let us focus for the moment on the standard binary linear classification task in which we encode the target class label as $t_i\in\{-1,1\}$ and the model parameter vector as ${\bf w}$. Letting $z_i= t_i {\bf w}^T {\bf x_i}$, 
we can define the logistic, hinge, and 0-1 loss as
\begin{eqnarray}
L_{log}(z_i)&=&\log[1+\exp( - z_i)] \label{eqn:logloss}\\
L_{hinge}(z_i) &=& \max(0, 1 - z_i) \label{eqn:hingeloss}\\
%
%
L_{01}(z_i) &=& \mathbb{I}[z_i \leq 0] \label{eqn:01loss}
\end{eqnarray}
where $\mathbb{I}[x]$ is the indicator function which takes the value of 1 when its argument is true and 0 when its argument is false. 
Of course, loss functions can be more complex, for example defined and learned through a linear combination of simpler basis loss functions \cite{vincent2002kernel}, but we focus on the widely used losses above for now. 
\cut{
\begin{figure}[ht!]
\centering
\includegraphics[width=6cm,height=4cm]{loss_functions.jpg} 
\caption{Four commonly used loss functions for the binary classification problem as a function of their input $z_i$: the 0-1 loss, \Rkh{$L_{01}(z_i)$}, the log logistic loss, \Rkh{$L_{\log}(z_i)$}, the hinge loss, \Rkh{$L_{hinge}(z_i)$}, and the squared loss, \Rkh{$L_{sq}(z_i)$}. }
\label{fig:log_hinge_01}
\end{figure}
}

\begin{figure}
\centering
\includegraphics[scale=.5]{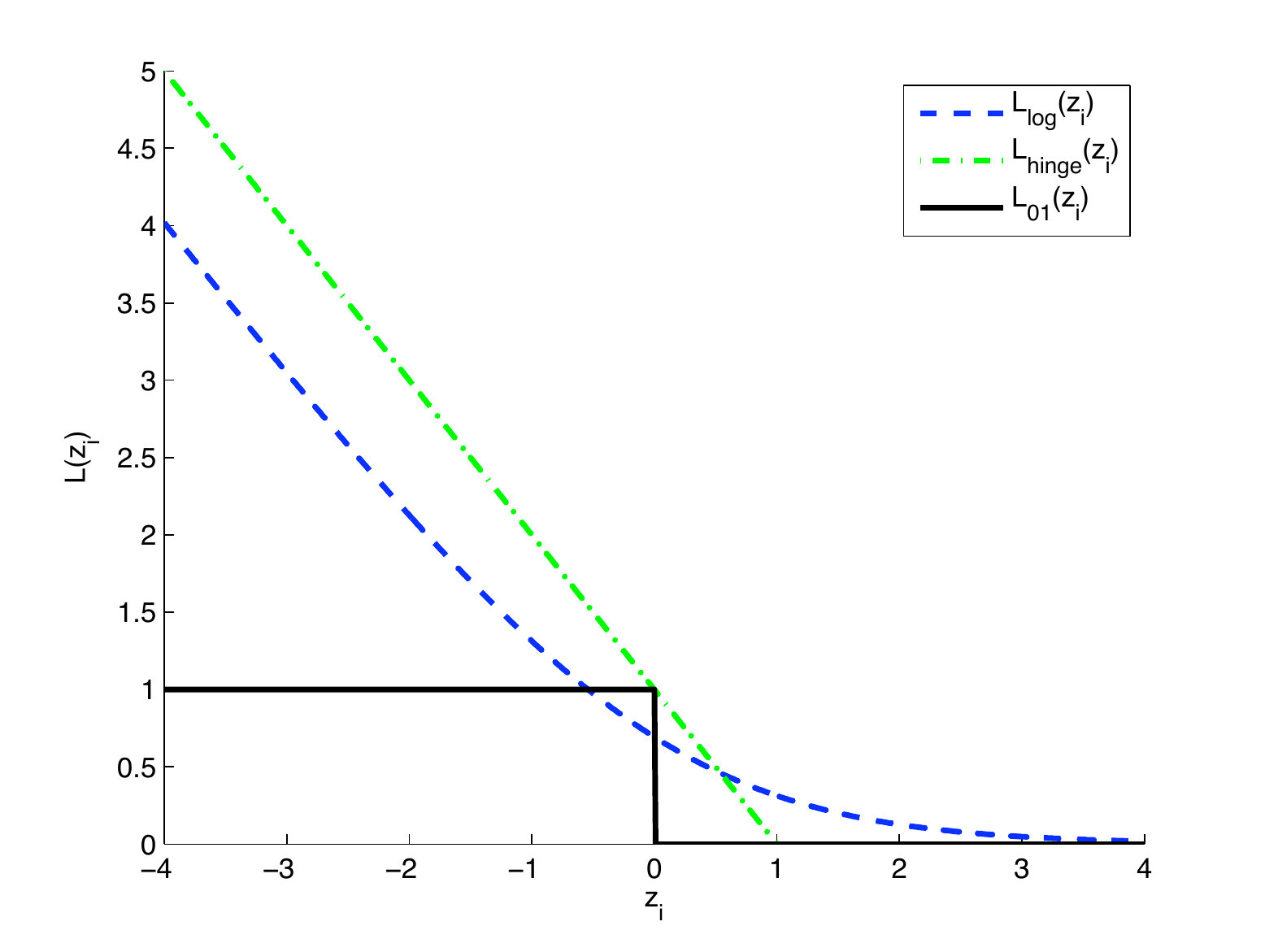}
\caption{Three widely used loss functions as a function of their input $z_i$: the log logistic loss, \Rkh{$L_{\log}(z_i)$}, the hinge loss, \Rkh{$L_{hinge}(z_i)$}, and the 0-1 loss, \Rkh{$L_{01}(z_i)$}. }
\label{fig:log_hinge_01}
\end{figure}

%
Different loss functions 
characterize the classification problem differently. 
The log logistic loss and the hinge loss are very similar in their shape, which can be verified from Figure \ref{fig:log_hinge_01}.
%
Logistic regression models involve optimizing the log logistic loss, while optimizing a hinge loss is the heart of Support Vector Machines (SVMs). 
While seemingly a sensible objective for a classification problem, empirical risk minimization with the 0-1 loss function is known to be an NP-hard problem \cite{feldman2012agnostic}.


Both the log logistic loss and the hinge loss are convex and therefore lead to optimization problems with a global minima. However, both the the log logistic loss and hinge loss penalize a model heavily when data points are classified incorrectly and are far away from the decision boundary. As can be seen in Figure \ref{fig:log_hinge_01} their penalties can be much more significant than the zero one loss.
%
%
%
%
The zero-one loss captures the intuitive goal of simply minimizing classification errors and
recent research has been directed to learning models using a smoothed zero-one loss approximation \cite{zhang2001text,nguyen2013algorithms}. 
%
%
%
Previous work has shown that both the hinge loss \cite{zhang2001text} and more recently the 0-1 loss \cite{nguyen2013algorithms} can be efficiently and effectively optimized directly using smooth approximations. The work in \cite{nguyen2013algorithms} also underscored the robustness advantages of the 0-1 loss to outliers. While the 0-1 loss is not convex, the current flurry of activity in the area of deep neural networks as well as the award winning work on 0-1 loss approximations in \cite{collobert2006trading} have highlighted numerous other advantages to the use of non-convex loss functions. 
%
%
In our work here, we are interested in constructing a probabilistically formulated smooth approximation to the 0-1 loss.

\begin{figure}
\centering
\includegraphics[scale=.5]{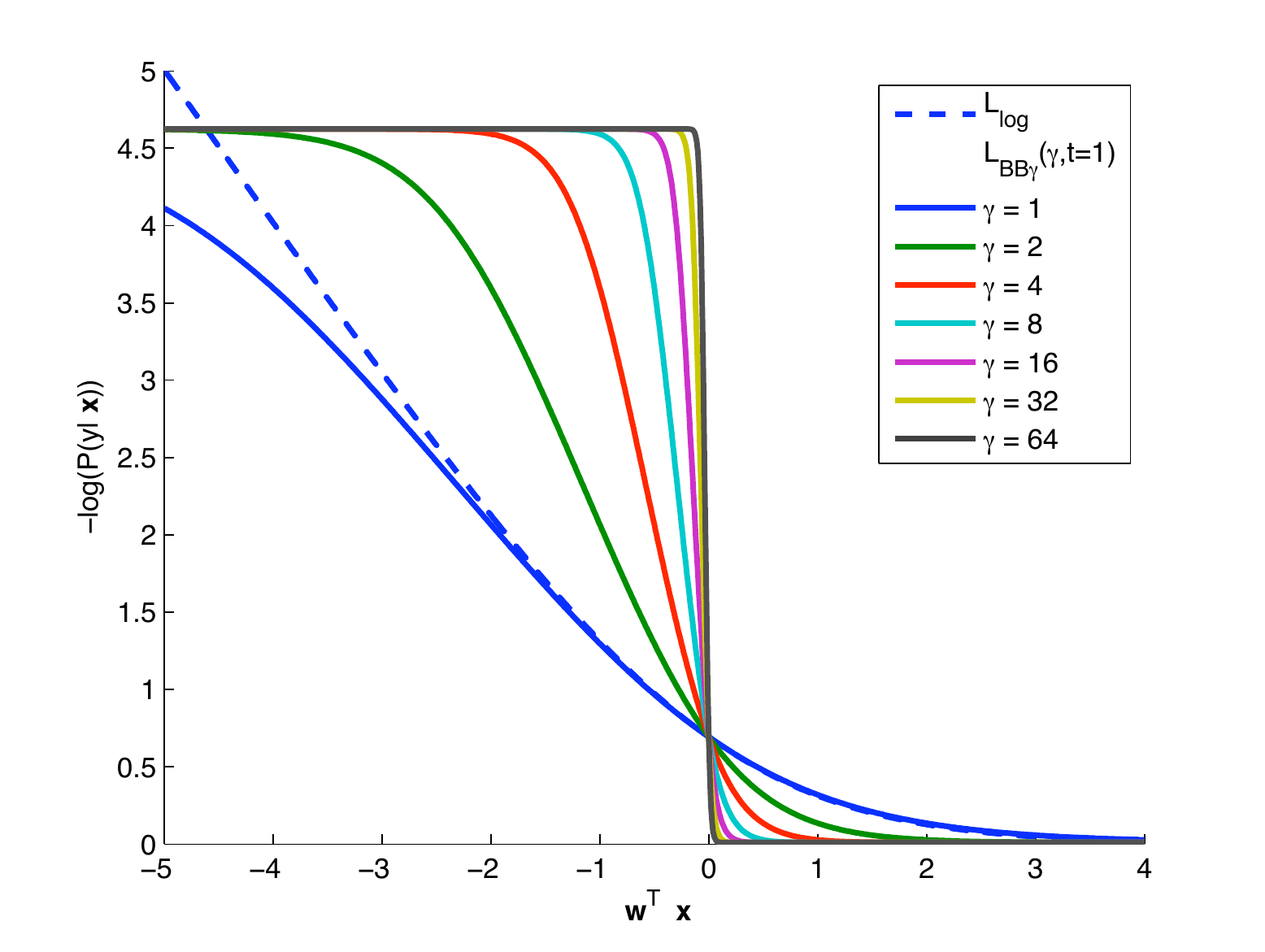}
\includegraphics[scale=.5]{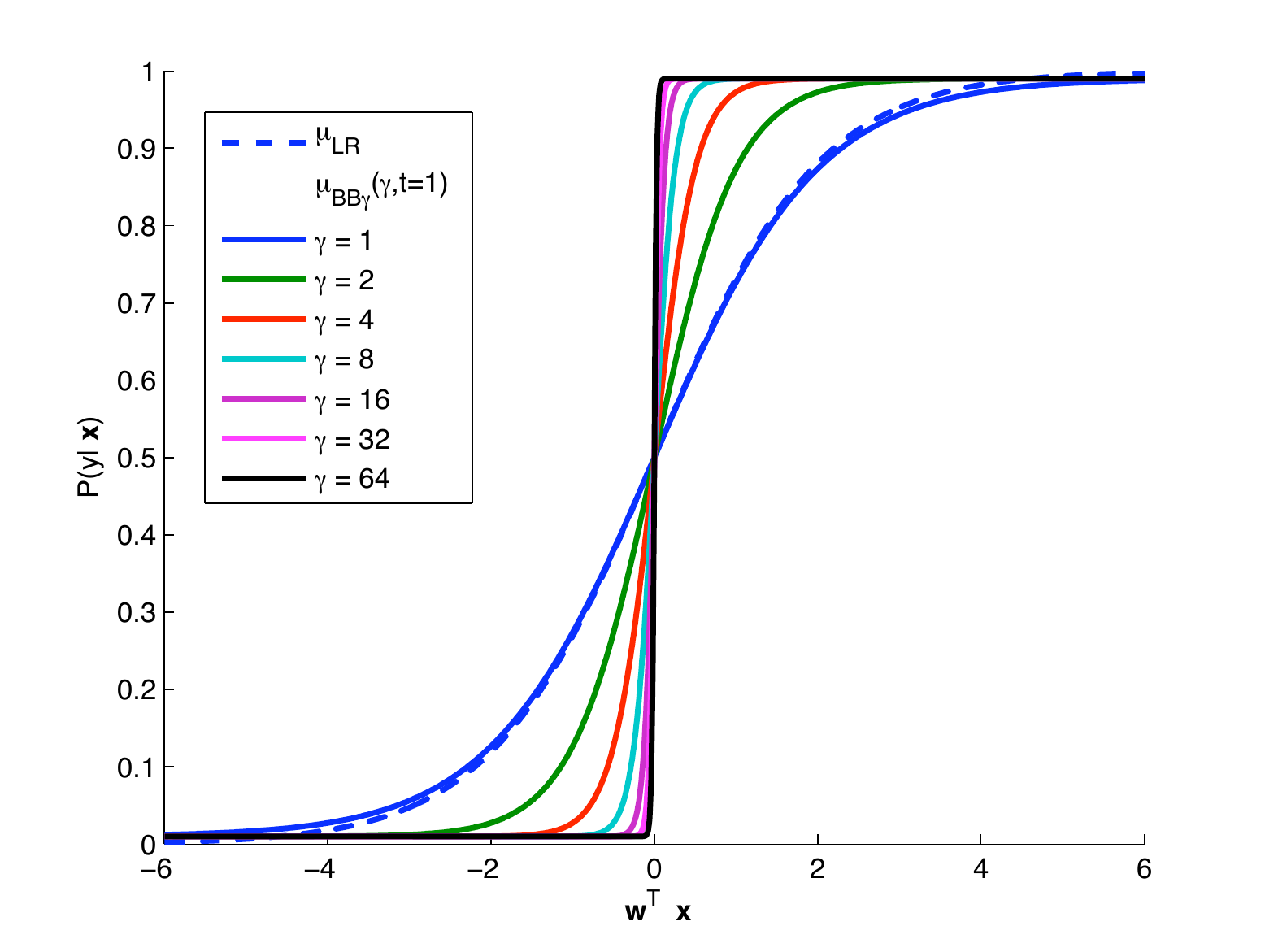}
\caption{(bottom panel) The probability, and (top panel) the corresponding negative log probability as a function of ${\bf w}^T{\bf x}$ for the log logistic loss compared with our generalized Beta-Bernoulli ($\mathcal{B} B\gamma$) model for different values of $\gamma$. We have used parameters $a=0.1$, $b=.98$, which corresponds to $\alpha=\beta=n/100$.
\Rkh{Here, $L_{\log}$ denotes the log logistic loss, $L_{\mathcal{B}B\gamma}$ denotes the Beta-Bernoulli loss, $\mu_{LR}$ denotes the Logistic Regression model (logistic sigmoid function), and $\mu_{\mathcal{B}B\gamma}$ denotes the generalized Beta-Bernoulli model}
}
\label{fig:prob-log}
\end{figure}

Let us first compare the widely used log logistic loss with the hinge loss and the 0-1 loss in a little more detail. 
\cut{
Let us also focus for the moment on binary linear classification in which we encode the target class as $t_i\in\{-1,1\}$ for feature vector ${\bf x}_i$ and we use a parameter vector ${\bf w}$. Letting $z_i= t_i {\bf w}^T {\bf x_i}$, the logistic, hinge and 0-1 losses can be expressed as 
\begin{eqnarray}
L_{log}(z_i)&=&\log[1+\exp( - z_i)] \label{eqn:logloss}\\
L_{hinge}(z_i) &=& \max(0, 1 - z_i) \label{eqn:hingeloss}\\
L_{01}(z_i) &=& \mathbb{I}[z_i \leq 0] \label{eqn:01loss}
\end{eqnarray}
where $\mathbb{I}[x]$ is the indicator function which takes the value of 1 when its argument is true and 0 when its argument is false. The overall loss is given by $\mathrm{L}=\sum_i^n L_{x}(z_i)$. We show these loss functions in Figure \ref{fig:log_hinge_01}.
\begin{figure}
\centering
\includegraphics[scale=.75]{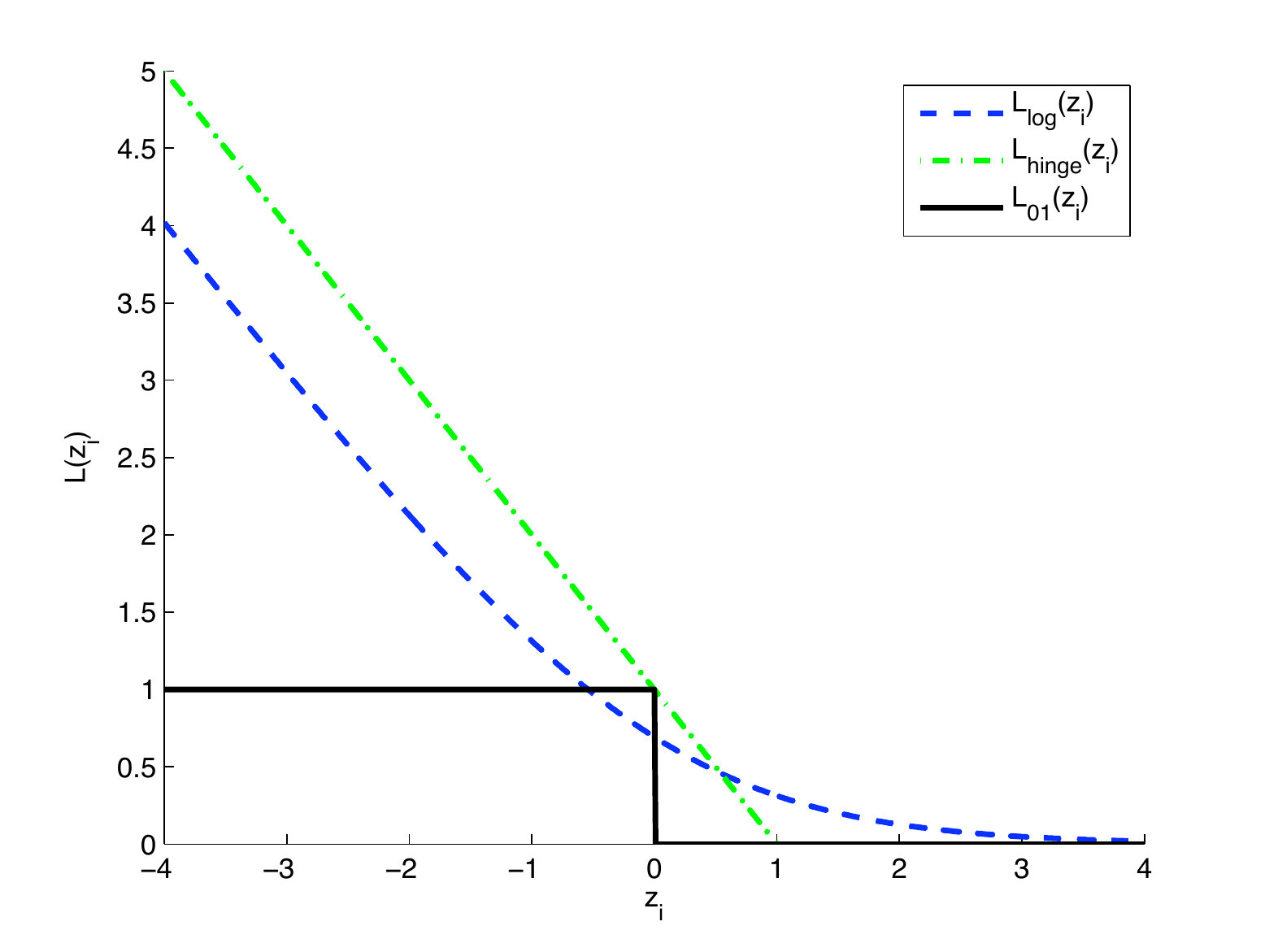}
\caption{Three widely used loss functions as a function of their input $z_i$: the log logistic loss, \Rkh{$L_{\log}(z_i)$}, the hinge loss, \Rkh{$L_{hinge}(z_i)$}, and the 0-1 loss, \Rkh{$L_{01}(z_i)$}. }
\label{fig:log_hinge_01}
\end{figure}

} 
The log logistic loss from the well known logistic regression model arises from the form of negative log likelihood defined by the model. 
More specifically, this logistic loss arises from a sigmoid function parametrizing probabilities and is easily recovered by re-arranging (\ref{eqn:logloss}) to obtain a probability model of the form $\pi(z_i)=(1+\exp(-z_i))^{-1}$. In our work here, we will take this familiar logistic function and we shall transform it to create a new functional form. The sequence of curves starting with the blue curve in Figure \ref{fig:prob-log} (top) give an intuitive visualization of the way in which we alter the traditional log logistic loss. We call our new loss function the generalized Beta-Bernoulli logistic loss and use the acronym $\mathcal{B} B\gamma$ when referring to it. We give it this name as it arises from the combined use of a Beta-Bernoulli distribution and a generalized logistic parametrization. 

We give the Bayesian motivations for our Beta-Bernoulli construction in section \ref{sec:theory}. To gain some additional intuitions about the effect of our construction from a practical perspective, consider the following analysis. When viewing the negative log likelihood of the traditional logistic regression parametrization as a loss function, one might pose the following question: (1) what alternative functional form for the underlying probability $\pi(z_i)$ would lead to a loss function exhibiting a plateau similar to the 0-1 loss for incorrectly classified examples? One might also pose a second question: (2) is it possible to construct a simple parametrization in which a single parameter controls the sharpness of the smooth approximation to the 0-1 loss? The intuition for an answer to the first question is that the traditional logistic parametrization converges to zero probability for small values of its argument. This in turn leads to a loss function that increases with a linear behaviour for small values of $z_i$ as shown in Figure \ref{fig:log_hinge_01}. 
In contrast, our new loss function is defined in such a way that for small values of $z_i$, the function will converge to a \emph{non-zero} probability. This effect manifests itself as the desired plateau, which can be seen clearly in the loss functions defined by our model in Figure \ref{fig:prob-log} (top). The answer to our second question is indeed yes; and more specifically, to control the sharpness of our approximation, we use a $\gamma$ factor reminiscent of a technique used in previous work which has created smooth approximations to the hinge loss \cite{zhang2001text} as well as smooth approximations of the 0-1 loss \cite{nguyen2013algorithms}. We show the intuitive effect of our construction for different increasing values of gamma in Figure \ref{fig:prob-log} and define it more formally below. 

To compare and contrast our loss function with other common loss functions such as those in equations (\ref{eqn:logloss}-\ref{eqn:01loss}) and others reviewed below,
 we express our loss here using $z_i$ and $\gamma$ as arguments. For $t=1$, the $\mathcal{B} B\gamma$ loss can be expressed as
\begin{equation}
L_{\mathcal{B} B\gamma}(z_i,\gamma)= -\log \big( a + b [1+\exp(-\gamma z_i)]^{-1} \big),
\label{eqn:BBg_pos}
\end{equation}
while for $t=-1$ it can be expressed as
\begin{equation}
L_{\mathcal{B} B\gamma}(z_i,\gamma)= -\log \big[ 1 - \big(a + b [1+\exp(\gamma z_i)]^{-1} \big) \big].
\label{eqn:BBg_neg}
\end{equation}
%
%
%
%
We show in section \ref{sec:theory} that the constants $a$ and $b$ have well defined interpretations in terms of the standard $\alpha$, $\beta$, and $n$ parameters of the Beta distribution. Their impact on our proposed generalized Beta-Bernoulli loss arise from applying a fuller Bayesian analysis to the formulation of a logistic function. 

The visualization of our proposed $\mathcal{B} B\gamma$ loss in Figure \ref{fig:prob-log} corresponds to the use of a weak non-informative prior such as $\alpha=1$ and $\beta=1$ and $n=100$. In Figure \ref{fig:prob-log}, we show the probability given by the model as a function of ${\bf w}^T{\bf x}$ at the right and the negative log probability or the loss on the left as $\gamma$ is varied over the integer powers in the interval $[0,10]$. We see that the logistic function transition becomes more abrupt as $\gamma$ increases. The loss function behaves like the usual logistic loss for $\gamma$ close to 1, but provides an increasingly more accurate smooth approximation to the zero one loss with larger values of $\gamma$. Intuitively, the location of the plateau of the smooth log logistic loss approximation on the y-axis is controlled by our choice of $\alpha$, $\beta$ and $n$. The effect of the weak uniform prior is to add a small minimum probability to the model, which can be imperceptible in terms of the impact on the sigmoid function log space, but leads to the plateau in the negative log loss function. By contrast, the use of a strong prior for the losses in Figure \ref{fig:BBstrong} (left) leads to minimum and maximum probabilities that can be much further from zero and one.

Our work makes a number of contributions which we enumerate here:
\textbf{(1)} The primary contribution of our work is a new probabilistically formulated approximation to the 0-1 loss based on a generalized logistic function and the use of the Beta-Bernoulli distribution. The result is a generalized sigmoid function in both probability and negative log probability space.  
\textbf{(2)} A second key contribution of our work is that we present and explore an adapted version of the optimization algorithm proposed in \cite{nguyen2013algorithms} in which we optimize the meta parameters of learning using validation sets. We present a series of experiments in which we optimize the $\mathcal{B} B\gamma$ loss using the basic algorithm from \cite{nguyen2013algorithms} and our modified version. For linear models, we show that our complete approach outperforms the widely used techniques of logistic regression and linear support vector machines. As expected, our experiments indicate that the relative performance of the approach further increases when noisy outliers are present in the data. \textbf{(3)} We go on to present a number of  experiments with larger scale data sets demonstrating that our method also outperforms widely used logistic regression and SVM techniques despite the fact that the underlying models involved are linear. \textbf{(4)} \Rkh{ We apply our model in a structured prediction task formulated for mining faces in Wikipedia biography pages. Our proposed method is well adapted to this setting and we and find that the improved probabilistic modeling capabilities of our approach yields improved results for visual information extraction through improved probabilistic structured prediction.} 
\textbf{(5)} 
We also show how this approach is also easily adapted to create a novel form of kernel logistic regression based on our generalized Beta-Bernoulli Logistic Regression (BBLR) framework. We find that the kernelized version of our method, Kernel BBLR (KBBLR) outperforms non-linear support vector machines. As expected, the $L_2$ regularized KBBLR does not yield sparse solutions; however, \textbf{(6)} since we have developed a robust method for optimizing a non-convex loss we propose and explore a novel non-convex sparsity encouraging prior based on a mixture of a Gaussian and a Laplacian. Sparse KBBLR typically yields sparser solutions than SVMs with comparable prediction performance, and the degree of sparsity scales much more favorably compared to SVMs .  

The remainder of this paper is structured as follows. In section \ref{sec:priorart}, we present a review of some relevant recent work in the area of 0-1 loss approximation. In section \ref{sec:theory}, we present the underlying Bayesian motivations for our proposed loss function. In section \ref{sec:algorithms}, we provide with the details of optimization and algorithms. In section \ref{sec:expts}, we present experimental results using protocols that both facilitate comparisons with prior work as well as evaluate our method on some large scale and structured prediction problems. We provide a final discussion and conclusions in section \ref{sec:Discuss_Conc}.

\section{Relevant Recent Work} 
\label{sec:priorart}

It has been shown in \cite{zhang2001text} that it is possible to define a generalized logistic loss and produce a smooth approximation to the hinge loss using the following formulation
\begin{eqnarray}
L_{glog}(t_i,{\bf x_i};{\bf w},\gamma) &=&\frac{1}{\gamma}\log[1+\exp( \gamma (1 - t_i {\bf w}^T {\bf x_i}))], \\
L_{glog}(z_i,\gamma) &=& \gamma^{-1}\log[1+\exp( \gamma (1 - z_i))],
\end{eqnarray}
such that $\lim_{\gamma \rightarrow \infty}L_{glog}=L_{hinge}$. We have achieved this approximation using a $\gamma$ factor and a shifted version of the usual logistic loss. We illustrate the way in which this construction can be used to approximate the hinge loss in Figure \ref{fig:gen_logloss} (left).

The maximum margin Bayesian network formulation in \cite{pernkopf2012maximum} also employs a smooth differentiable hinge loss inspired by the Huber loss, having a similar shape to $min[1,z_i]$. 
The sparse probabilistic classifier approach in \cite{herault2007sparse} 
truncates the logistic loss leading to a sparse kernel logistic regression models. 
\cite{perez2003empirical} proposed a technique for learning support vector classifiers based on arbitrary loss functions composed of using the combination of a hyperbolic tangent loss function and a polynomial loss function.

Other recent work \cite{nguyen2013algorithms} has created a smooth approximation to the 0-1 loss by directly defining the loss as a modified sigmoid. They used the following function
\begin{eqnarray}
L_{sig}(t_i,{\bf x_i};{\bf w},\gamma) &=&\frac{1}{1+\exp(\gamma t_i {\bf w}^T {\bf x_i}) }, \\
L_{sig}(z_i,\gamma) &=&[1+\exp(\gamma z_i)]^{-1}.
\end{eqnarray}
In a way similar to the smooth approximation to the hinge loss, here $\lim_{\gamma \rightarrow \infty}L_{sig}=L_{01}$. We illustrate the way in which this construction approximates the 0-1 loss in Figure \ref{fig:gen_logloss} (right). 
%

\begin{figure}[ht]
\vskip 0.2in
\begin{center}
\centerline{\includegraphics[scale=.35]{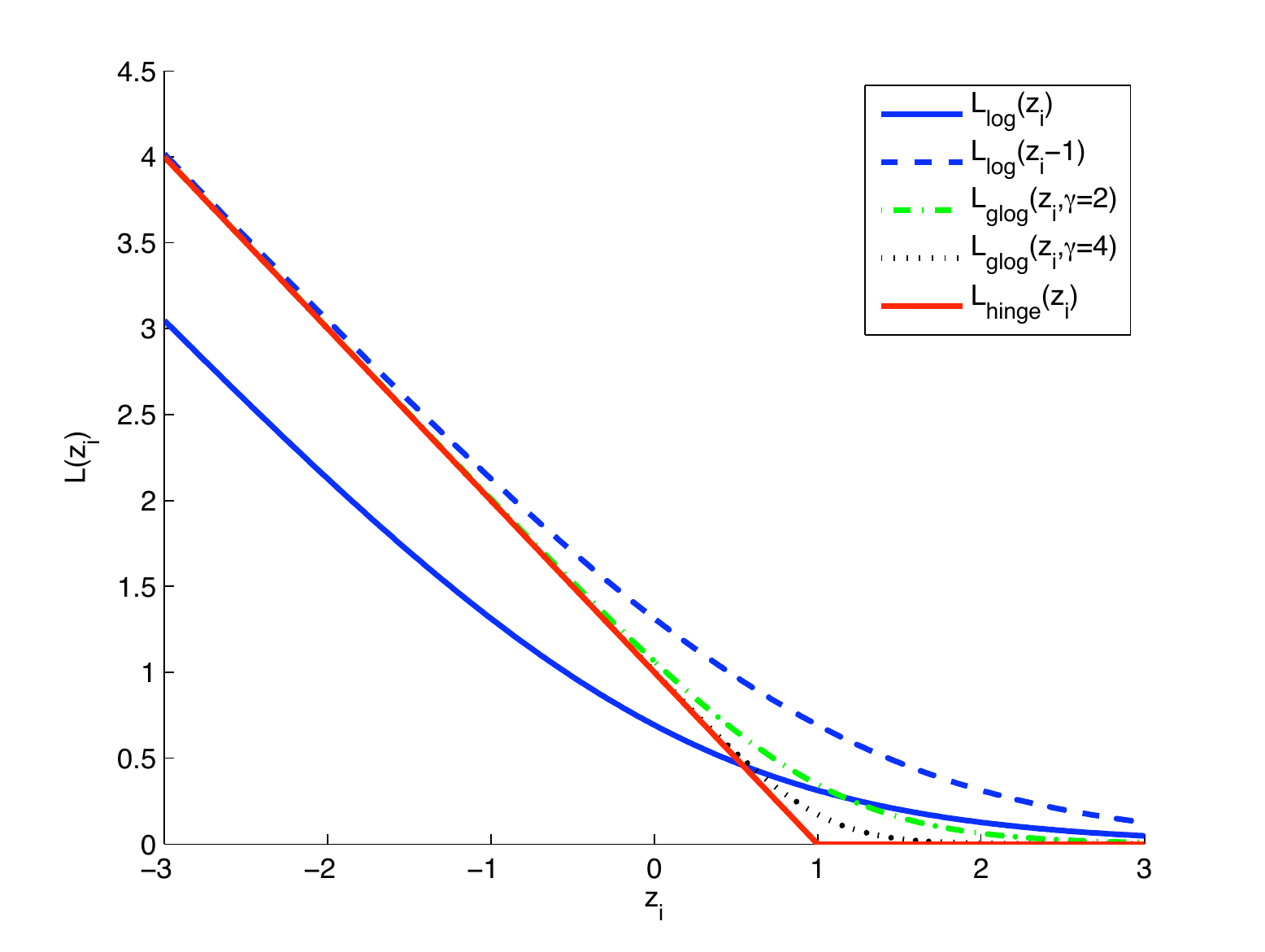}
\includegraphics[scale=.35]{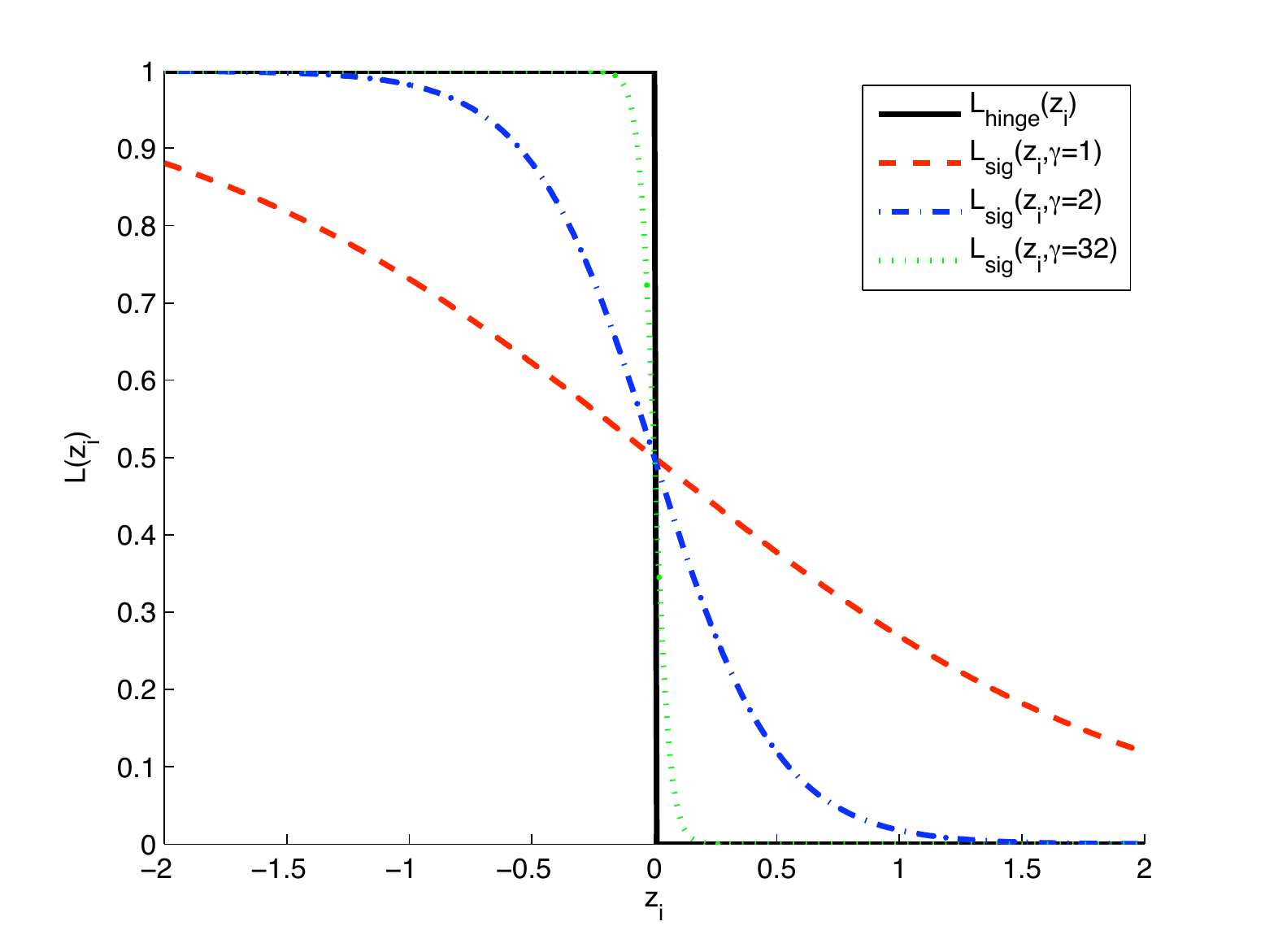}}
\caption{({\bf left}) The way in which the generalized log logistic loss, $L_{glog}$ proposed in \cite{zhang2001text} can approximate the hinge loss, $L_{hinge}$ through translating the log logistic loss, $L_{\log}$ then increasing the $\gamma$ factor. We show here the curves for $\gamma=2$ and $\gamma=4$. %
({\bf right}) The way in a sigmoid function is used in \cite{nguyen2013algorithms} to directly approximate the 0-1 loss, $L_{01}$. The approach also uses a similar $\gamma$ factor to \cite{zhang2001text} and we show $\gamma=1,2$ and $32$.}
\label{fig:gen_logloss}
\end{center}
\vskip -0.2in
\end{figure}

Another important aspect of \cite{nguyen2013algorithms} is that they compared a variety of algorithms for directly optimizing the 0-1 loss with a novel algorithm for optimizing the sigmoid loss, $L_{sig}(z_i,\gamma)$. They call their \emph{algorithm} Smooth 0–1 Loss Approximation (SLA) for smooth loss approximation. The compared direct 0-1 loss optimization algorithms are: (1) a Branch and Bound (BnB) \cite{land1960automatic} technique, (2) a Prioritized Combinatorial Search (PCS) technique and  (3) an algorithm referred to as a Combinatorial Search Approximation (CSA), both of which are presented in more detail in \cite{nguyen2013algorithms}.  They compared these methods with the use of their SLA algorithm to optimize the sigmoidal approximation to the 0-1 loss.

To evaluate and compare the quality of the non-convex optimization results produced by the BnB, PCS and CSA, with their SLA algorithm for the sigmoid loss, \cite{nguyen2013algorithms} also presents training set errors for a number of standard evaluation datasets. We provide an excerpt of their results in Table \ref{tab:SLA_losses} as we will perform similar comparisons in our experimental work. These results indicated that the SLA algorithm consistently yielded superior performance at finding a good minima to the underlying non-convex problem. Furthermore, in \cite{nguyen2013algorithms}, they also provide an analysis of the run-time performance for each of the algorithms. 
Their experiments indicated that the SLA technique was significantly faster than the alternative algorithms for non-convex optimization. Based on these results we build upon the SLA approach in our work here.

\begin{table}[ht!]
\begin{center}
\caption{An excerpt from \cite{nguyen2013algorithms} of the \BRkh{0-1 loss performance}{total 0-1 loss} for a variety of algorithms on some standard datasets. The 0-1 loss for logistic regression (LR) and a linear support vector machine (SVM) are also provided for reference. }
\label{tab:SLA_losses}
\begin{small}
\begin{tabular}{|c||c|c||c|c|c|c|}
\hline
      & LR & SVM & PCS & CSA & BnB & SLA \\ \hline \hline
Breast & 19 & 18 & 19 & 13 & 10 & 13 \\ \hline
Heart & 39 & 39 & 33 & 31 & 25 & 27  \\ \hline
Liver & 99 & 99 & 91 & 91 & 95 & 89  \\ \hline
Pima & 166 & 166 & 159 & 157 & 161 & 156  \\ \hline \hline
\BRkh{Total}{Sum} & 323 & 322 & 302 & 292 & 291 & 285  \\ \hline
\end{tabular}
\end{small}
\end{center}
\end{table}

\cut{
\begin{table}
\begin{center}
\caption{An excerpt from \cite{nguyen2013algorithms} for the running times associated with the results summarized in Table \ref{tab:SLA_losses}. Times are given in seconds. NA indicates that the corresponding algorithm could not find a better solution than its given initial solution given a maximum running time. }
\label{tab:time}
\begin{small}
\begin{tabular}{|c||c|c||c|c|c|c|}
\hline
      & LR & SVM & PCS & CSA & BnB & SLA \\ \hline \hline
Breast & 0.05 & 0.03 & NA & 161.64 & 3.59 & 1.13 \\ \hline
Heart & 0.03 & 0.02 & 1.24 & 126.52 & 63.56 & 0.77  \\ \hline
Liver & 0.01 & 0.01 & 97.07 & 16.11 & 0.17 & 0.39  \\ \hline
Pima & 0.04 & 0.03 & 63.30 & 157.38 & 89.89 & 0.89  \\ \hline \hline
\end{tabular}
\end{small}
\end{center}
\end{table}
}

\begin{figure}[ht!]
\centering
\includegraphics[width=\columnwidth]{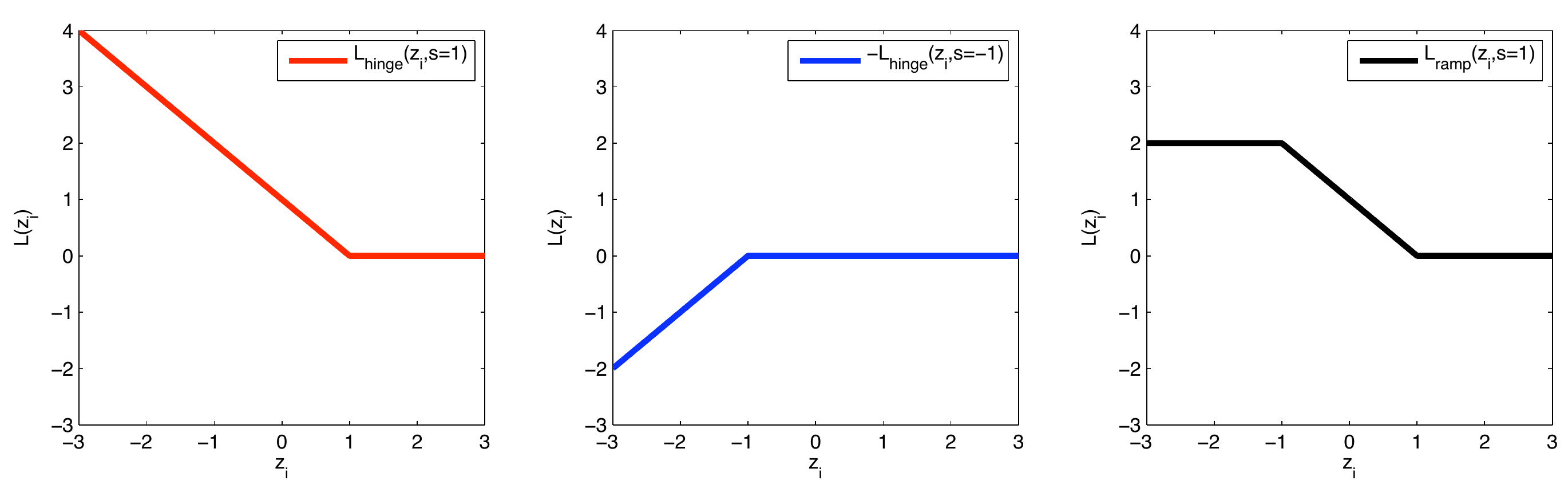}
\caption{The way in which shifted hinge losses are combined in \cite{collobert2006trading} to produce the ramp loss, $L_{ramp}$. The usual hinge loss (left), $L_{hinge}$ is combined with the negative, shifted hinge loss, $L_{hinge}(z_i,s=-1)$ (middle), to produce $L_{ramp}$ (right).}
\label{fig:ramp_loss}
\end{figure}

The award winning work of \cite{collobert2006trading} produced an approximation to the 0-1 loss by creating a ramp loss, $L_{ramp}$, obtained by combining the traditional hinge loss with a shifted and inverted hinge loss as illustrated in Figure \ref{fig:ramp_loss}. They showed how to optimize the ramp loss using the Concave-Convex Procedure (CCCP) of \cite{yuille2003concave} and that this yields faster training times compared to traditional SVMs. Other more recent work has proposed an alternative online SVM learning algorithm for the ramp loss \cite{ertekin2011nonconvex}. \cite{wu2007robust} explored a similar ramp loss which they refer to as a robust truncated hinge loss. More recent work \cite{cotter2013learning} has explored a similar ramp like construction which they refer to as the slant loss. Interestingly, the ramp loss formulation has also been generalized to structured predictions \cite{do2008tighter,gimpel2012structured}.

\cut{
\begin{figure}[ht!]
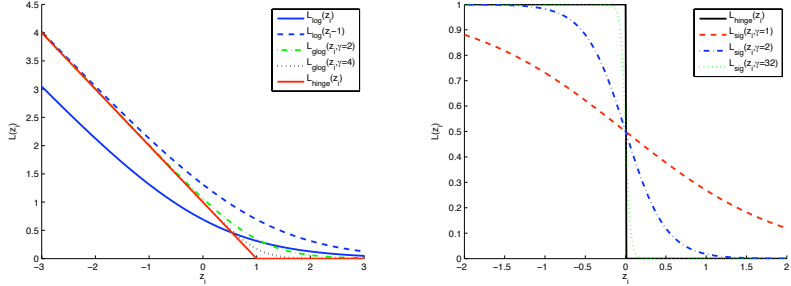

\centering
\includegraphics[scale=.25]{gamma_trick_better_2.pdf}
\includegraphics[scale=.25]{sigmoid_loss.pdf}
\caption{({\bf left}) The way in which the generalized log logistic loss, $L_{glog}$ proposed in \cite{zhang2001text} can approximate the hinge loss, $L_{hinge}$ through translating the log logistic loss, $L_{\log}$ then increasing the $\gamma$ factor. We show here the curves for $\gamma=2$ and $\gamma=4$. %
({\bf right}) The way in a sigmoid function is used in \cite{nguyen2013algorithms} to directly approximate the 0-1 loss, $L_{01}$. The approach also uses a similar $\gamma$ factor to \cite{zhang2001text} and we show $\gamma=1,2$ and $32$.
%
}
\label{fig:gen_logloss}
\end{figure}
} 

\cut{
\begin{figure}[ht!] 
\centering
\includegraphics[scale=.25]{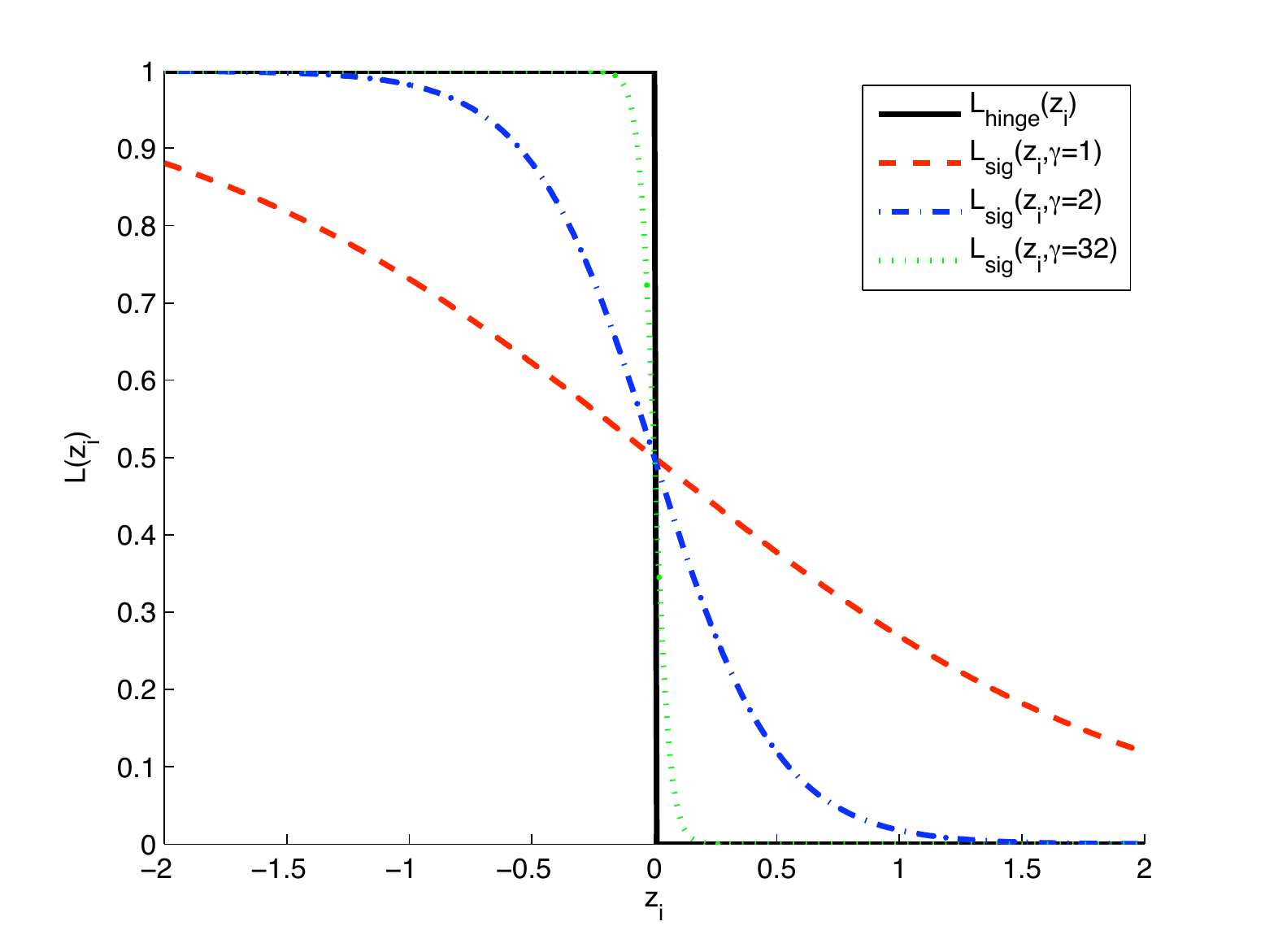}
\caption{The way in a sigmoid function is used in \cite{nguyen2013algorithms} to directly approximate the 0-1 loss, $L_{01}$. The approach also uses a similar $\gamma$ factor to \cite{zhang2001text} and we show $\gamma=1,2$ and $32$. 
\Rkh{
$L_{sig}$ denotes the sigmoid loss, and $L_{hinge}$ denotes the hinge loss.
}
}
\label{fig:sigmoid_loss}
\end{figure}
} 

Although the smoothed zero-one loss captured much attention recently, we can find older references to similar research. There has been the activity of using zero-one loss like functional losses in machine learning, specially by the boosting \cite{mason1999boosting} and neural network \cite{vincent2004modeles} communities.
Vincent \cite{vincent2004modeles} analyzes that the loss defined through a functional of the hyperbolic tangent,  $1-\tanh (z_i)$, is more robust as it doesn't penalize the outliers too excessively compared to other $\{$ log logistic loss, hinge loss, and squared loss $\}$ loss functions. %
This loss has interesting properties of both being continuous and with zero-one loss like properties.
A variant of this loss has been used in boosting algorithms 
\cite{mason1999boosting}. Other work \cite{vincent2004modeles} has also shown that a hyperbolic tangent parametrized squared error loss, $(0.65-\tanh(z_i))^2$, transforms the squared error loss to behave more like the $1-\tanh(z)$, hyperbolic tangent loss. 

We shall see below how it is also possible to integrate our novel smooth loss formulation into models for structured prediction. In this way our work is similar to that of \cite{gimpel2012structured} which explored the use of the ramp loss of \cite{collobert2006trading} in the context of structured prediction for machine translation.

\section{Our Approach: Generalized Beta-Bernoulli Logistic Classification}
\label{sec:theory}

We now derive a novel form of logistic regression based on formulating a generalized sigmoid function arising from an underlying Bernoulli model with a Beta prior. We also use a $\gamma$ scaling factor to increase the sharpness of our approximation. 
Consider first the traditional and widely used formulation of logistic regression which can be derived from a probabilistic model based on the Bernoulli distribution. The Bernoulli probabilistic model has the form:
\begin{equation}
P(y|\theta)=\theta^y(1-\theta)^{(1-y)},
\end{equation}
where $y\in\{0,1\}$ \Rkh{ is the class label, and $\theta$ is the parameter of the model.} 
The Bernoulli distribution can be re-expressed in standard exponential family form as
\begin{equation}
P(y|\theta)=\exp \Big\{ \log\Big(\frac{\theta}{1-\theta}\Big)y + \log(1-\theta) \Big\},
\end{equation}
where the natural parameter $\eta$ is given by
\begin{equation}
\eta = \log\Big(\frac{\theta}{1-\theta}\Big)
\end{equation}
In traditional logistic regression, we let the natural parameter $\eta = {\bf w}^T {\bf x}$, 
which leads to a model where $\theta=\theta_{ML}$ in which the following parametrization is used
\begin{equation}
\theta_{ML} =\mu_{ML}(\bf w, \bf x) =\frac{1}{1+\exp(-\eta)} 
   = \frac{1}{1+\exp(-{\bf w}^T {\bf x})}
\label{eqn:sigmoid}
\end{equation}
The conjugate distribution to the Bernoulli is the Beta distribution
\begin{equation}
\mathcal{B}\textmd{eta}(\theta|\alpha,\beta) = \frac{1}{\mathrm{B}(\alpha,\beta)}\theta^{\alpha-1}(1-\theta)^{\beta-1}
\end{equation}
where $\alpha$ and $\beta$ have the intuitive interpretation as the equivalent pseudo counts for observations for the two classes of the model and $\mathrm{B}(\alpha,\beta)$ is the beta function.
When we use the Beta distribution as the prior over the parameters of the Bernoulli distribution, the posterior mean of the Beta-Bernoulli model is easily computed due to the fact that the posterior is also a Beta distribution.
%
This property also leads to an intuitive form for the posterior mean or expected value $\theta_{\mathcal{B} B}$ in a Beta-Bernoulli model, which consists of a simple weighted average of the prior mean $\theta_{\mathcal{B}}$ and the traditional maximum likelihood estimate, $\theta_{ML}$, such that
\begin{equation}
\label{eqn:bblr_ML}
\theta_{\mathcal{B} B} = w \theta_{\mathcal{B}} + (1-w)\theta_{ML},
\end{equation}
where
\begin{equation}
w = \frac{\alpha+\beta}{\alpha+\beta+n}, \: \textmd{and} \:\: \theta_{\mathcal{B}} = \Big(\frac{\alpha}{\alpha+\beta}\Big), \nonumber
\end{equation}
and where $n$ is the number of examples used to estimate $\theta_{ML}$.
Consider now the task of making a prediction using a Beta posterior and the predictive distribution. It is easy to show that the mean or expected value of the posterior predictive distribution is equivalent to plugging the posterior mean parameters of the Beta distribution into the Bernoulli distribution, $\textmd{Ber}(y|\theta)$, i.e.
\begin{equation}
p(y |\mathcal{D}) = \int_0^1 p(y|\theta)p(\theta|\mathcal{D})d\theta=\textmd{Ber}(y|\theta_{\mathcal{B} B}).
\end{equation}

%

%

Given these observations, we thus propose here to replace the traditional sigmoidal function used in logistic regression with the function given by the posterior mean of the Beta-Bernoulli model such that
%
%
%
%
\begin{equation}
\mu_{\mathcal{B} B}({\bf w}, {\bf x})= w\Big(\frac{\alpha}{\alpha+\beta}\Big) + (1-w)\mu_{ML}({\bf w}, {\bf x})
\end{equation}
%




Further, to increase our model's ability to approximate the zero one loss, we shall also use a generalized form of the Beta-Bernoulli model above where we set the natural parameter of $\mu_{ML}$ so that $\eta = \gamma {\bf w}^T {\bf x}$. This leads to our complete model based on a generalized Beta-Bernoulli formulation
\begin{equation}
\mu_{\mathcal{B} B\gamma}({\bf w}, {\bf x})= w\Big(\frac{\alpha}{\alpha+\beta}\Big) + (1-w)\frac{1}{1+\exp(-\gamma{\bf w}^T {\bf x})}.
\end{equation}
It is useful to remind the reader at this point that we have used the Beta-Bernoulli construction to define our \emph{function}, not to define a prior over the parameter of a random variable as is frequently done with the Beta distribution.  
Furthermore, in traditional Bayesian approaches to logistic regression, a prior is placed on the parameters $w$ and used for MAP parameter estimation or more fully Bayesian methods in which one integrates over the uncertainty in the parameters.

In our formulation here, we have placed a  prior on the \emph{function} $\mu_{ML}({\bf w},{\bf x})$ as is commonly done with Gaussian processes. Our approach might be seen as a pragmatic alternative to working with the fully Bayesian posterior distributions over functions given data, $p(f|\mathcal{D})$. The more fully Bayesian procedure would be to use the posterior predictive distribution to make predictions using
\begin{equation}
p(y_*|x_*,\mathcal{D}) = \int p(y_*|f,x_*)p(f|\mathcal{D})df.
\end{equation}


Let us consider again the negative log logistic loss function defined by our generalized Beta-Bernoulli formulation where we let $z = \textbf{w}^T\textbf{x}$ and we use our $y\in{\{0,1\}}$ encoding for class labels. For $y=1$ this leads to  
\begin{equation}
-\log p(y=1|z)=
-\log \Bigg[w \theta_{\beta} + \frac{(1-w)}{1 + \exp(-\gamma z)} \Bigg], 
\end{equation}
while for the case when $y=0$, the negative log probability is simply
\begin{equation}
-\log p(y=0|z)=
-\log \Bigg(1 -\Bigg[ w \theta_{\beta} + \frac{(1-w)}{1 + \exp(-\gamma z)} \Bigg] \Bigg) 
\end{equation}
where $w \theta_{\beta}=a$ and $(1-w)=b$ for the formulation of the corresponding loss given earlier in equations (\ref{eqn:BBg_pos}) and (\ref{eqn:BBg_neg}). 

\begin{figure}[ht!]
\centering
\includegraphics[scale=.35]{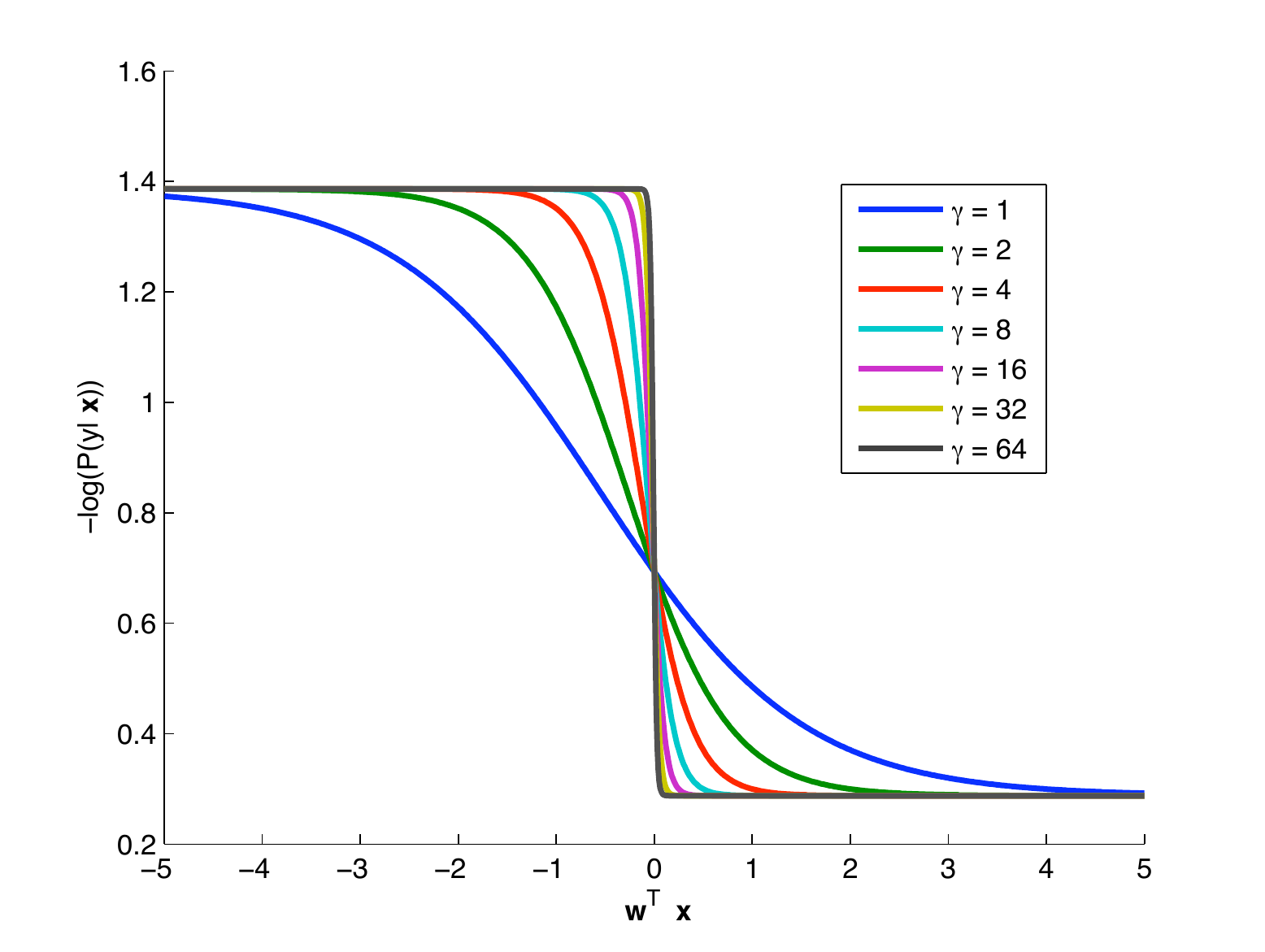}
\includegraphics[scale=0.35]{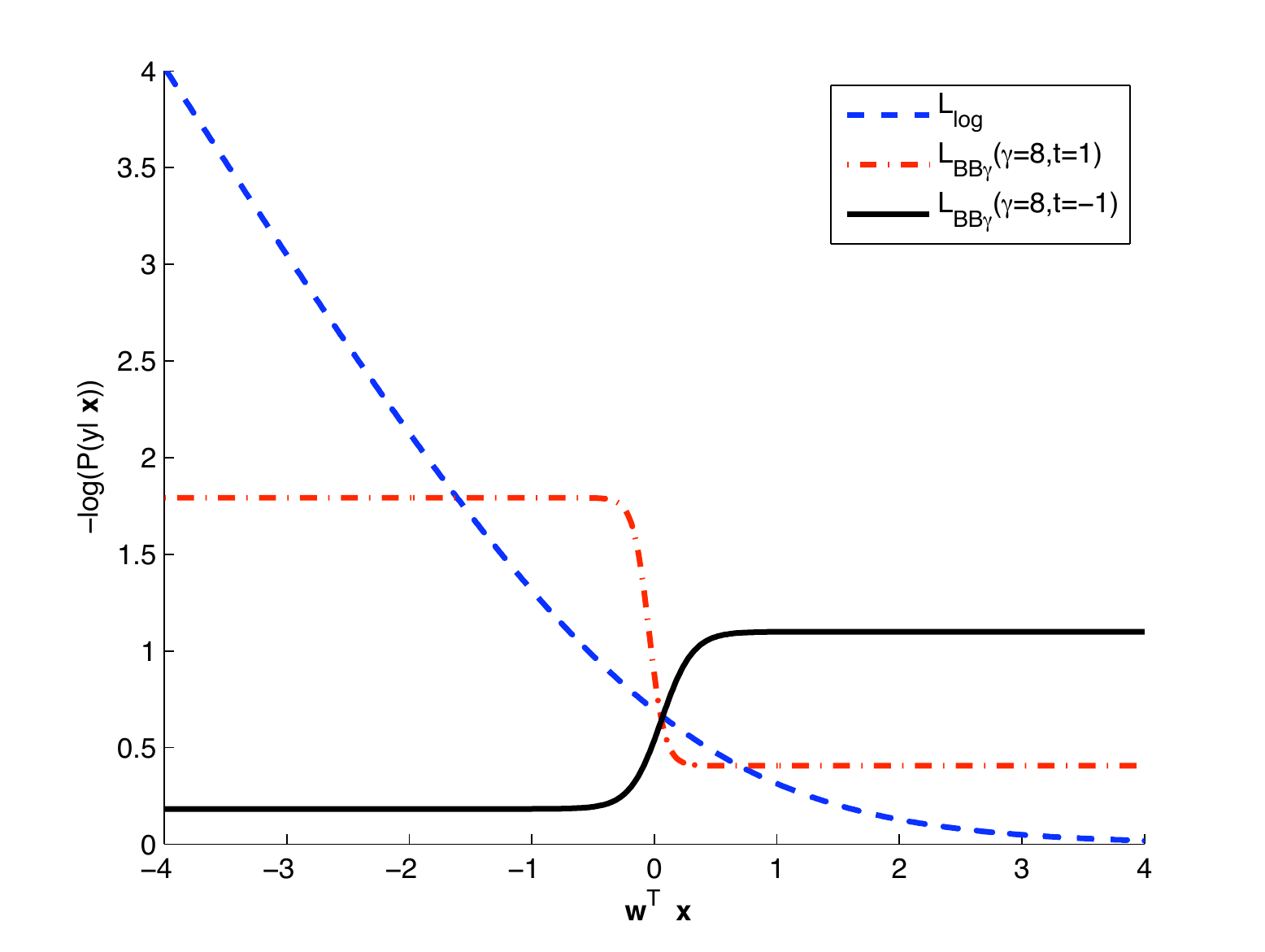}
\caption{({\bf left}) The $\mathcal{B} B\gamma$ loss, or the negative log probability for $t=1$ as a function of ${\bf w}^T{\bf x}$ under our generalized Beta-Bernoulli model for different values of $\gamma$. We have used parameters $a=1/4$ and $b=1/2$, which corresponds to $\alpha=\beta=n/2$. 
({\bf right})
The $\mathcal{B} B\gamma$ loss also permits asymmetric loss functions. We show here the negative log probability for both $t=1$ and $t=-1$ as a function of ${\bf w}^T{\bf x}$ with $\gamma=8$. This loss corresponds to $\alpha=n,\beta=2n$. We also give the log logistic loss as a point of reference. 
\Rkh{Here, $L_{\log}(z_i)$ denotes the log logistic loss, and $L_{\mathcal{B}B\gamma}(z_i)$ denotes the Beta-Bernoulli loss.}
}
%
\label{fig:BBstrong}
\end{figure}

In Figure \ref{fig:prob-log} we showed how setting this scalar parameter $\gamma$ to larger values, i.e $\gg 1$ allows our generalized Beta-Bernoulli model to more closely approximate the zero one loss. 
%
We show the $\mathcal{B} B\gamma$ loss with $a=1/4$ and $b=1/2$ in Figure \ref{fig:BBstrong} (left) which corresponds to a stronger Beta prior and as we can see, this leads to an approximation with a range of values that are even closer to the 0-1 loss.
As one might imagine, with a little analysis of the form and asymptotics of this function, one can also see that for given a setting of $\alpha=\beta$ and $n$, a corresponding scaling factor $s$ and linear translation $c$ can be found so as to transform the range of the loss into the interval $[0, 1]$ such that $\lim_{\gamma \rightarrow \infty}s(L_{\mathcal{B} B\gamma} -c) = L_{01}$. However, when $\alpha \neq \beta$ as shown in Figure \ref{fig:BBstrong} (right), the loss function is asymmetric and in the limit of large gamma this corresponds to different losses for true positives, false positives, true negatives and false negatives. For these and other reasons we believe that this formulation has many attractive and useful properties.

\cut{
\begin{figure}[ht!] 
\includegraphics[scale=0.25]{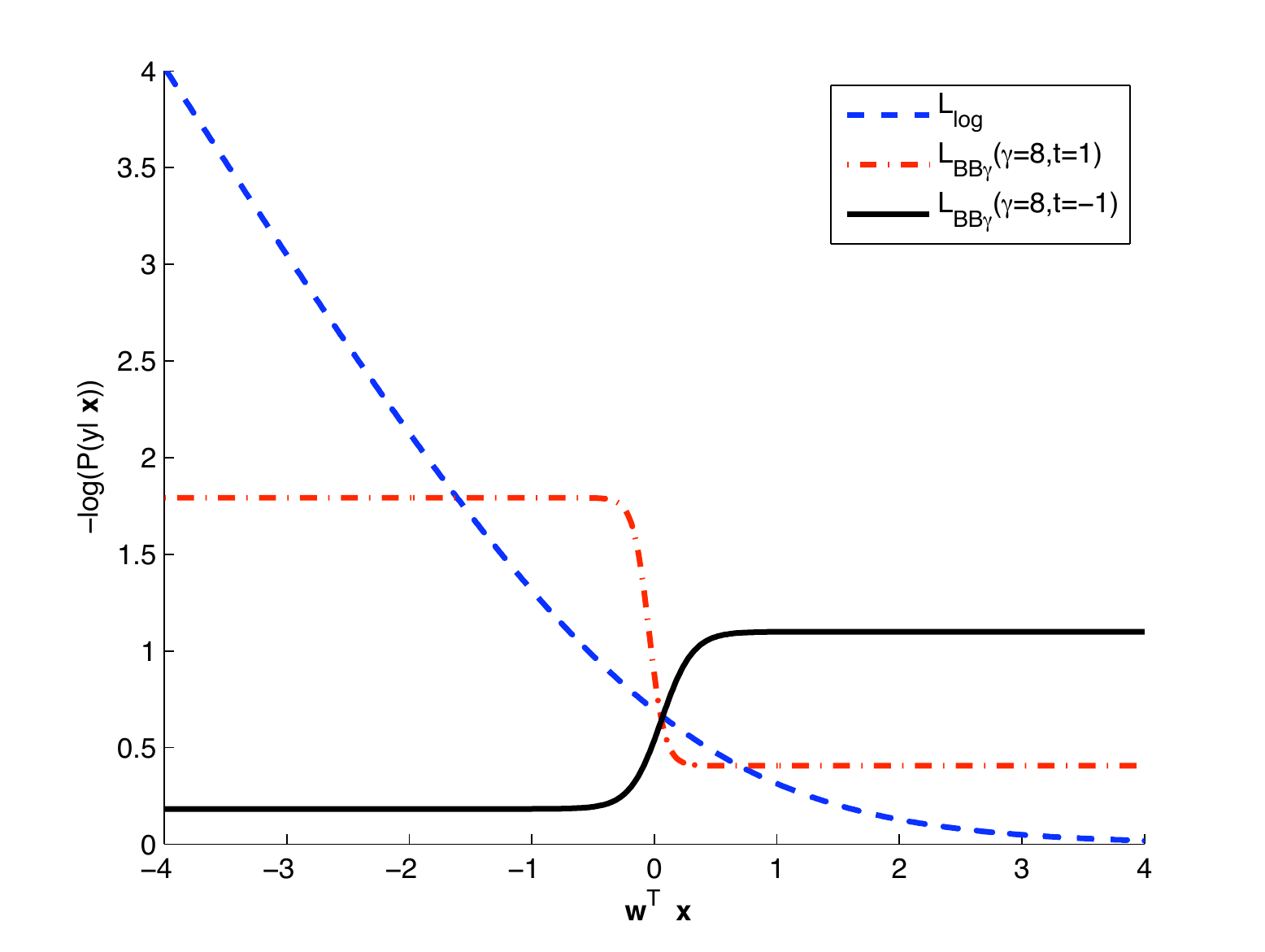}
\caption{The $\mathcal{B} B\gamma$ loss also permits asymmetric loss functions. We show here the negative log probability for both $t=1$ and $t=-1$ as a function of ${\bf w}^T{\bf x}$ with $\gamma=8$. This loss corresponds to $\alpha=n,\beta=2n$. We also give the log logistic loss as a point of reference. 
\Rkh{Here, $L_{\log}(z_i)$ denotes the log logistic loss, and $L_{\mathcal{B}B\gamma}(z_i)$ denotes the Beta-Bernoulli loss.}
}
\label{fig:BBasym}
\end{figure}
}

\subsection{Parameter Estimation and Gradients}

We now turn to the problem of estimating the parameters ${\bf w}$, given data in the form of $D=\{y_i,{\bf x}_i\}, i=1,\ldots,n$, using our model. As we have defined a probabilistic model, as usual we shall simply write the probability defined by our model then optimize the parameters via maximizing the log probability or minimizing the negative log probability. As we shall discuss in more detail in section $\ref{sec:algorithms}$, we use a modified form of the SLA optimization algorithm in which we slowly increase $\gamma$ and interleave gradient descent steps with coordinate descent implemented as a grid search. For the gradient descent part of the optimization we shall need the gradients of our loss function and we therefore give them below.

Consider first the usual formulation of the conditional probability used in logistic regression
\begin{equation}
P(\{y_i\}|\{{\bf x}_i\},{\bf w})
=\prod_{i=1}^n \mu_{i}^{y_i}(1-\mu_{i})^{(1-{y_i})},
\end{equation}
here in place of the usual $\mu_i$, in our generalized Beta-Bernoulli formulation we now have \BRkh{$\mu_i = \mu_{\beta B}(\eta({\bf w}, {\bf x}_i,\gamma))$ where $\eta_i=\gamma {\bf w}^T{\bf x}$}{$\mu_i = \mu_{\beta B}(\eta_i)$ where $\eta_i=\gamma {\bf w}^T{\bf x}_i$}. Given a data set $D$ consisting of label and feature vector pairs, this yields a log-likelihood given by
\begin{equation}
{\cal L} = \log P(D|{\bf w})  \\
=\sum_{i=1}^n \big( {y_i} \log \mu_{i} + (1-{y_i}) \log (1 - \mu_{i}) \big)
\label{eqn:bblr_ll}
\end{equation}
where the gradient of this function is given by
\begin{equation}
\frac{d {\cal L}}{d{\bf w}}  
=\sum_{i=1}^n \Big( \frac{y_i}{\mu_i} -\frac{1-y_i}{1-\mu_i} \Big) \frac{d\mu_i}{d\eta_i} \frac{d \eta_i}{d {\bf w}} 
\label{eqn:gd_bblr}
\end{equation}
%
%
%
with
\begin{equation}
\frac{d\mu_i}{d\eta_i}= 
(1-w)\frac{\exp(-\eta_i)}{(1+\exp(-\eta_i))^2}
\: \: \textmd{and} \: \: 
\frac{d \eta_i}{d {\bf w}} = \gamma {\bf x}.
\end{equation}
\cut{
such that
\begin{equation}
\frac{dL}{d{\bf w}}  
=\gamma(1-w) \sum_{i=1}^n {\bf x}_i (y_i-\mu_i) \Big(\frac{\mu_{ML}}{\mu_i}\Big)\Big(\frac{1-\mu_{ML}}{1-\mu_i}\Big)
\end{equation}
%
%
Using equations 3 and 5, we can derive
\begin{align}
\mu_i &= w\Big(\frac{\alpha}{\alpha+\beta}\Big) + (1-w)(1+\exp(-\eta_i))^{-1} \\ \nonumber
\frac{d\mu_i}{d\eta_i}&=(1-w)\frac{\exp(-\eta_i)}{(1+\exp(-\eta_i))^2}\\ \nonumber
&=(1-w)\frac{\exp(-\eta_i)}{(1+2\exp(-\eta_i)+\exp(-2\eta_i))}\\ \nonumber
&=(1-w)\frac{1}{\exp(\eta_i)+2+\exp(-\eta_i)}\\ \nonumber
%
\end{align}
}
\cut{
Taking the derivative with respect to $\theta_{\beta}$, yields
\begin{equation}
\frac{d{\cal L}}{d\theta_{\beta}} = w \sum_{i=1}^n 
\frac{y_i-\mu_i}{\mu_i(1-\mu_i)}.
\end{equation}
The derivative for $w$ is
\begin{equation}
\frac{d{\cal L}}{dw} = \sum_{i=1}^n 
\frac{y_i-\mu_i}{\mu_i(1-\mu_i)}(\theta_{\beta} - \mu_i).
\end{equation}
} 

\subsection{Some Asymptotic Analysis}

As we have stated at the beginning of our discussion on parameter estimation, at the end of our optimization we will have a model with a large $\gamma$. With a sufficiently large $\gamma$ all predictions will be given their maximum or minimum probabilities possible under the $\beta B \gamma$ model. Defining the $t=1$ class as the positive class, if we set the maximum probability under the model equal to the True Positive Rate (TPR) (e.g. on training and/or validation data) and the maximum probability for the negative class equal to the True Negative Rate (TNR) we have
\begin{eqnarray}
w\theta_{\beta}+(1-w)&=&TPR, \\
1-w\theta_{B}&=&TNR,
\end{eqnarray}
which allows us to conclude that this would equivalently correspond to setting
\begin{eqnarray}
w&=&2-(TNR+TPR), \\
\theta_{B}&=&\frac{1-TNR}{2-(TNR+TPR)}.
\end{eqnarray}
This analysis gives us a good idea of the expected behavior of the model if we optimize $w$ and $\theta_{B}$ on a training set. It also suggests that an even better strategy for tuning $w$ and $\theta_{B}$ would be to use a validation set. 

%
%

\subsection{Learning hyper-parameters}
\label{sec:w_ThB}
We have provided an asymptotic analysis of the expected values for $w$ and $\theta_B$ in the previous section. In the experiment section, we provide BBLR results for using asymptotic values of these two parameters along with cross-validated values for other hyper-parameters $\{\gamma, \lambda \}$, where $\lambda $ is the regularization parameter described in Section \ref{sec:algorithms}. It is however also possible to learn these hyper-parameters using the training set, validation set or both. Below, we provide partial-derivatives of likelihood function (\ref{eqn:bblr_ll}) for these hyper-parameters. 
\cut{
\begin{equation}
{\cal L}(D|{w,\theta_B})  
=\sum_{i=1}^n {y_i} \log \mu_{i} + (1-{y_i}) \log (1 - \mu_{i})
\label{eqn:BLL_no_l2}
\end{equation}
} 
%

%

\begin{equation}
\frac{d{\cal L}}{d{w}}  
=\sum_{i=1}^n \Big( \frac{y_i}{\mu_i} -\frac{1-y_i}{1-\mu_i} \Big) \frac{d\mu_i}{d\\w} 
\label{eqn:gd_w}
\end{equation}

with
\begin{equation}
\frac{d\mu_i}{d\\w}=  
\theta_B-\frac{1}{1+\exp(-\gamma{\bf w}^T{\bf x})}
\end{equation}
The partial-derivatives with respect to $\theta_B$ and $\gamma$ are as follows

%
\begin{equation}
\frac{d{\cal L}}{d{\theta_B}}  
=w\sum_{i=1}^n \Big( \frac{y_i}{\mu_i} -\frac{1-y_i}{1-\mu_i} \Big)  
\label{eqn:gd_w}
\end{equation}
\begin{equation}
\frac{d {\cal L}}{d{\gamma}}  
=\sum_{i=1}^n \Big( \frac{y_i}{\mu_i} -\frac{1-y_i}{1-\mu_i} \Big) \frac{d\mu_i}{d\eta_i} \frac{d \eta_i}{d {\gamma}} 
\label{eqn:gd_bblr_gamma}
\end{equation}

with
\begin{equation}
\frac{d\mu_i}{d\eta_i}= 
(1-w)\frac{\exp(-\eta_i)}{(1+\exp(-\eta_i))^2}
\: \: \textmd{and} \: \: 
\frac{d \eta_i}{d {\gamma}} = {\bf w}^T{\bf x}.
\end{equation}

\subsection{Kernel Beta-Bernoulli Classification}
\label{sec:bb_klr}

It is possible to transform the traditional logistic regression technique discussed above into a kernel logistic regression (KLR)
by replacing the linear discriminant function, $\eta=\textbf{w}^T\textbf{x}$, with 
\begin{equation}
\eta=f(\textbf{a},\textbf{x})=\displaystyle\sum_{j=1}^{N}a_j K(\textbf{x},\textbf{x}_j),
\end{equation}
where $K(\textbf{x},\textbf{x}_j)$ is a kernel function and $j$ is used as an index in the sum over all $N$ training examples. 

To create our generalized Beta-Bernoulli KLR model we take a similar path; however, in this case we let $\eta=\gamma f(\textbf{a},\textbf{x})$. Thus, our Kernel Beta-Bernoulli model can be written as:
\begin{equation}
\mu_{K\beta B}({\bf a},{\bf x})= w\Big(\frac{\alpha}{\alpha+\beta}\Big) + \frac{(1-w)}{1 + \exp \big(-\gamma f(\textbf{a},{\bf x})\big)}.
\label{eqn:KBBLR_mu}
\end{equation}
If we write $f(\textbf{a},\textbf{x})= \textbf{a}^T \textbf{k}(\textbf{x})$, where $\textbf{k}(\textbf{x})$ is a vector of kernel values, then the gradient of the corresponding KBBLR log likelihood obtained by setting $\mu_i=\mu_{K\beta B}({\bf a},{\bf x})$ in (\ref{eqn:bblr_ll}) is
%
\begin{equation}
\frac{dL}{d{\bf a}}  
=\gamma(1-w) \sum_{i=1}^n \textbf{k}(\textbf{x}_i) \Big( \frac{y_i}{\mu_i} -\frac{1-y_i}{1-\mu_i} \Big)\frac{\exp(-\eta_i)}{(1+\exp(-\eta_i))^2}.
\end{equation}

\section{Optimization and Algorithms}
\label{sec:algorithms}

As we have discussed in \BRkh{our literature review}{the relevant recent work section above}, the work of \cite{nguyen2013algorithms} has shown that their SLA algorithm applied to $L_{sig}(z_i,\gamma)$ outperformed a number of other techniques in terms of both true 0-1 loss minimization performance and run time. As our generalized Beta-Bernoulli loss, $L_{BB\gamma}(z_i,\gamma)$ is another type of smooth approximation to the 0-1 loss, we therefore use a variation of their SLA algorithm to optimize the $BB\gamma$ loss. 
%
%
%
Recall that if one compares our generalized Beta-Bernoulli logistic loss with the directly defined sigmoidal loss used in the SLA work of \cite{nguyen2013algorithms}, it becomes apparent that the BBLR formulation has three additional hyper-parameters, $\{\alpha, \beta, w\}$. These additional parameters control the locations of the plateaus of our function and these plateaus have well defined interpretations in terms of probabilities. In contrast, the plateaus of the sigmoidal loss in \cite{nguyen2013algorithms} are located at zero and one. Additionally, in practise one is interested in optimizing the regularized loss, where some form of prior or regularization is used for parameters. In our experiments here, we follow the widely used practice of using a Gaussian prior for parameters.
The corresponding regularized loss arising from the negative log likelihood with the additional $L_2$ regularization term gives us our complete objective function
\begin{equation}
L(D|{\bf w})  
=-\sum_{i=1}^n \big( {y_i} \log \mu_{i} + (1-{y_i}) \log (1 - \mu_{i}) \big) +\frac{\lambda}{2}\|{\bf w}\|^2,
\label{eqn:gd_bblr_rg_l2}
\end{equation}
where the parameter $\lambda$ controls the strength of the regularization.
With these additional hyper-parameters $\{\alpha, \beta, w,\lambda\}$, the original SLA algorithm is not directly applicable to our formulation. However, if we hold these hyper-parameters fixed, we are able to use the general idea of their approach and perform a Modified SLA optimization as given in Algorithms 1 and 2. In our experiments below, we use that strategy in the BBLR$^{1,2,3}$ series of experiments.
To deal with the issue of how to jointly learn weights ${\bf w}$ as well as hyper-parameters $w$, $\alpha$, $\beta$, $\lambda$ and $\gamma$; in our BBLR$^{4}$ series of experiments we learn these hyper-parameters by gradient descent on the training set. More precisely, we learn $w$ and $\theta_{B}$ (as opposed to learning $\alpha$, $\beta$) as this permit the parameters to be easily re-parametrized so that they both lie within $[0,1]$.  

Very importantly, our initial experiments indicated that the basic SLA formulation required considerable hand tuning of learning parameters for each new data set. This was the case even using the simplest smooth loss function without the additional degrees of freedom afforded by our formulation. This led us to develop a meta-optimization procedure for learning algorithm parameters. The BBLR$^{3,4}$ series of experiments below use this learning meta-parameter optimization procedure. Our initial and formal experiments here indicate that this meta-optimization of learning parameters is in fact essential in practice.  We therefore present it in more detail below. 
%
%
%
%
%
%
\subsection{Our SLA Algorithm Meta-optimization (SLAM)}

Here we present our meta-optimization extension and various other modifications to the SLA approach of \cite{nguyen2013algorithms}. 
The SLA algorithm proposed in \cite{nguyen2013algorithms} can be decomposed into two different parts; an outer loop that initializes a model then enters a loop in which one slowly increases the $\gamma$ factor of their sigmoidal loss, repeatedly calling an algorithm they refer to as \emph{Range Optimization for SLA} or \emph{Gradient Descent in Range}. The Range Optimization part consists of two stages. \emph{Stage 1} is a standard gradient descent optimization with a decreasing learning rate (using the new $\gamma$ factor). \emph{Stage 2} probes each parameter $w_i$ in a radius $R$ using a one dimensional grid search to determine if the loss can be further reduced, thus implementing a coordinate descent on a set of grid points.  We provide a slightly modified form of the outer loop of their algorithm in Algorithm \ref{Algorithm:SLA_outer} where we have expressed the initial parameters given to the model, ${\bf w}_0$ as explicit parameters given to the algorithm. In their approach they hard code the initial parameter estimates as the result of an SVM run on their data. 
We provide a compressed version of their inner Range optimization technique in Algorithm \ref{Algorithm:SLA_range}.  

\begin{algorithm}[htb!]
\caption{Modified SLA - Initialization and outer loop}
\label{Algorithm:SLA_outer}
\begin{algorithmic}[1]
\Require Training data ${\bf X},{\bf t}$, and initial weights ${\bf w}_0$, 
and
\Statex \textbf{constants}: $R_0, \epsilon_{S_0},\gamma_{MIN},\gamma_{MAX}, r_\gamma, r_R, r_{\epsilon}$
\Ensure ${\bf w}^*$, estimated weights minimizing 0-1 loss.
\Function{Find-SLA-Solution}{${\bf X}$,${\bf t}$,${\bf w}_0$}
\State ${\bf w} \gets {\bf w}_0$
\State $R \gets R_0$
\State $\epsilon_S \gets \epsilon_{S_0}$
\State $\gamma \gets \gamma_{MIN}$
\While{$\gamma \leq \gamma_{MAX}$}
\State ${\bf w}^* \gets$ \Call{Grad-Desc-in-Range}{${\bf w}^*, {\bf X}, {\bf t}, \gamma,R,\epsilon_S$}
\State $\gamma \gets r_\gamma \gamma$
\State $R \gets r_R R$
\State $\epsilon_S \gets r_{\epsilon}\epsilon_S$
\EndWhile \label{end_of_while}
\State \textbf{return} ${\bf w}^*$
\EndFunction
\end{algorithmic}

\end{algorithm}

\begin{algorithm}[htb!]

\caption{Range Optimization for SLA}
\label{Algorithm:SLA_range}
\begin{algorithmic}[1]
\Require ${\bf w}, {\bf X}, {\bf t}, \gamma$, radius $R$, step size $\epsilon_S$
\Ensure Updated estimate for ${\bf w}^*$,  minimizing 0-1 loss.
\Function{Grad-Desc-In-Range}{${\bf w}, {\bf X}, {\bf t}, \gamma$, $R$, $\epsilon_S$}
\Repeat 
\Statex \hspace{1cm}$\rhd$\textmd{ Stage 1: Find local minimum}
\State ${\bf w}^* \gets$ \Call{Vanilla-Grad-Desc}{${\bf w}$}
\Statex \hspace{1cm}$\rhd$\textmd{ Stage 2: Probe each dimension in a radius R}
\Statex \hspace{1cm}$\rhd$\textmd{ to escape local minimum (if possible)}
%
%
\For{$i=1\ldots d$}\Comment{For each dimension, $w_i$}
\For{$step \in \{\epsilon_S,-\epsilon_S,2\epsilon_S,-2\epsilon_S,\dots,R,-R\}$}
\State  ${\bf w} \gets {\bf w}^*$
\State $w_i \gets w_i + step$
\If{$L_{\gamma}({\bf w}^*)-L_{\gamma}({\bf w}) \geq \epsilon_L$}
\State \textbf{break} \Comment{Goto step 3}
\EndIf
\EndFor 
\EndFor
\Until{$L_{\gamma}({\bf w}^*)-L_{\gamma}({\bf w}) < \epsilon_L$}
\State \textbf{return} ${\bf w}^*$
\EndFunction
\end{algorithmic}

\end{algorithm}

The first minor difference between the SLA optimization algorithm of \cite{nguyen2013algorithms} and our extension to it are the selection of the initial ${\bf w}_0$ that the SLA algorithm starts optimizing. While the original SLA algorithm uses the SVM solution as its initial solution, ${\bf w}_0$, our modified SLA algorithm uses the  $\gamma$ and $\lambda$ obtained from experiments using a validation set defined within the training data to initialize ${\bf w}_0$ for the gradient based optimization technique which will start from ${\bf w}={\bf 0}$. 
The idea here is to search for the best $\gamma$ and $\lambda$ that produces a reasonable solution of ${\bf w}$ that the SLA algorithm will start with, 
where $\lambda$ is the weight associated with the Gaussian prior leading to L2 penalty added to (\ref{eqn:bblr_ll}). 

Our meta-optimization procedure consists of the following.
We use the suggested values in the original SLA algorithm \cite{nguyen2013algorithms} for the parameters $r_R, R_0,r_\epsilon$, and $\epsilon_{S_0}$. For the others, we use a cross validation run using the same modified SLA algorithm to fine-tune algorithm parameters.

%

%
\cut{
\Rkh{
As we have mentioned earlier, the initial ${\bf w} \leftarrow {\bf w_0}$ (line 2, algorithm 1) is selected through a cross-validation run and a gradient based optimization algorithm. The idea here is to search for the best $\gamma$ and $\lambda$ that produce a reasonable solution of ${\bf w}$ that the SLA algorithm will start with, 
where $\lambda$ is the weight associated with the L2 regularizer added to (\ref{eqn:gd_bblr}). 
The modified likelihood is 
\begin{equation*}
L(D|{\bf w})  
=\sum_{i=1}^n {y_i} \log \mu_{i} + (1-{y_i}) \log (1 - \mu_{i})-\frac{\lambda}{2}\|{\bf w}\|^2
\label{eqn:gd_bblr_rg_l2}
\end{equation*}
}
} 
\cut{
and the corresponding gradient of this function is given by
\begin{equation}
\frac{dL}{d{\bf w}}  
=\sum_{i=1}^n \Big( \frac{y_i}{\mu_i} -\frac{1-y_i}{1-\mu_i} \Big) \frac{d\mu_i}{d\eta_i} \frac{d \eta_i}{d {\bf w}} - \lambda{\bf w} 
\label{eqn:gd_bblr_rg}
\end{equation}
} 
%
%
%
\Rkh{
%
Parameter $r_{\gamma}$ is chosen through a grid search, while $\gamma_{MIN}$ and $\gamma_{MAX}$ are chosen by a bracket search algorithm. 
%
%
} 
%
In our experience, these model parameters change from problem (dataset) to problem, and hence must be fine-tuned for the best results. 
%

\section{Experimental Setup and Results}
\label{sec:expts}

%
Below, we present results for three different groups of benchmark problems: (1) a selection from the University of California Irvine (UCI) repository, (2) some larger and higher dimensionality text processing tasks from the LibSVM evaluation archive \footnote{http://www.csie.ntu.edu.tw/~cjlin/libsvmtools/datasets/binary.html}, and (3) the product review sentiment prediction datasets used in  \cite{dredze2008confidence}. We then present results on a structured prediction problem formulated for the task of visual information extraction from Wikipedia biography pages. Finally we explore the kernelized version of our classifier.

In all experiments, unless otherwise stated, we use a Gaussian prior on parameters leading to an $L_2$ penalty term.  We explore four experimental configurations for our BBLR approach: {\bf (1)} BBLR$^1$, where we use our modified SLA algorithm 
with the following BBLR parameters held fixed : $\alpha=\beta=1$ and $n=100$. This corresponds to a minor modification to the traditional negative log logistic loss, but yields a probabilistically well defined smooth sigmoid shaped loss (ex. as we have seen in Figure \ref{fig:prob-log}); {\bf (2)} BBLR$^2$, where we use values for $\alpha, \beta$ and $n$ corresponding to the empirical counts of positives, negatives and the total number of examples from the training set, which corresponds to a simplistic heuristic, partially justified by Bayesian reasoning; {\bf (3)} BBLR$^3$ in which an outer meta-optimization of learning parameters is performed on top of (2), ie SLAM, and {\bf (4)} BBLR$^4$ in which the outer meta-optimization of learning parameters is performed, and the hyper-parameters $w$, $\theta_B$, $\lambda$, and $\gamma$ are optimized by gradient descent using the training set, with $w$ and $\theta_B$ initialized using the values given by our asymptotic analysis using a hard threshold for classifications. 
At each iteration of this optimization step, as  parameters $\{w,\theta_B,\lambda,\gamma\}$ get updated, the complementary SLAM hyper-parameters, $\gamma_R$, $\gamma_{MIN}$, $\gamma_{MAX}$ are adjusted/redefined by using the same meta-optimization procedure (SLAM) and using a subset of the training data as a validation set.

Consequently, models produced by the BBLR$^3$ series of experiments explore the ability of our improved SLA learning parameter meta-optimization method (SLAM) to effectively minimize a smooth approximation to the zero one loss. While the BBLR$^4$ series of experiments delve the deepest into the ability of our BBLR formulation and SLAM optimization to more accurately make probabilistic predictions.

\subsection{Binary Classification Tasks}

\subsubsection{Experiments with UCI Benchmarks}

We evaluate our technique on the following datasets from the University of California Irvine (UCI) Machine Learning Repository \cite{Bache+Lichman_2013}: Breast, Heart, Liver and Pima. We use these datasets in part so as to compare directly with results in \cite{nguyen2013algorithms}, to understand the behaviour of our novel logistic function formulation and to explore the behavior of our learning parameter optimization procedure.
Table \ref{tab:uci_details} shows some brief details of these databases. 
%

\begin{table}[htb!]
\begin{center}
\caption{Standard UCI benchmark datasets used for our experiments.  }
\label{tab:uci_details}
\begin{tabular}{|c||p{2.45cm}|p{2.2cm}|p{4.7cm}|}
\hline
Dataset & \# Examples & \# Dimensions & Description \\ \hline \hline 
Breast & 683 & 10 &  Breast Cancer Diagnosis \cite{mangasarian1995breast} \\ \hline
Heart & 270 & 13 & Statlog \\ \hline
Liver & 345 & 6 & Liver Disorders \\ \hline
Pima & 768 & 8 & Pima Indians Diabetes \\ \hline \hline
\end{tabular}
\end{center}
\end{table}


To facilitate comparisons with previous results presented  \cite{nguyen2013algorithms} such as those summarized in Table \ref{tab:01_all} of our literature review in Section \ref{sec:priorart}, we provide a small set of initial experiments here following their experimental protocols.  
In our experiments here we compare our BBLRs with the following models: our own L2 Logistic Regression (LR) implementation, a linear SVM - using the same implementation (liblinear) that was used in \cite{nguyen2013algorithms}, and the optimization of the sigmoid loss, $L_{sig}(z_i,\gamma)$ of \cite{nguyen2013algorithms} using the SLA algorithm and the code distributed on the web site associated with \cite{nguyen2013algorithms} (indicated by SLA in our tables). 
\cut{
In our BBLR implementations, we used the same SLA algorithm with the following modifications ---
%
rather than using the result of an SVM as the initialization, we use the result of a grid search over values of $\gamma$ and our Gaussian prior over parameters from a simple gradient descent run with our model. The free parameters of the LR and SVM models, used in the above and in the subsequent experiments are chosen through cross validations. 
} 

Despite the fact that we used the code distributed on the website associated with \cite{nguyen2013algorithms} we found that the SLA algorithm applied to their sigmoid loss, $L_{sig}(z_i,\gamma)$ gave errors that are slightly higher than those given in \cite{nguyen2013algorithms}. We use the term SLA in Table \ref{tab:01_all} and subsequent tables to denote experiments performed using both the sigmoidal loss explored in \cite{nguyen2013algorithms} and their algorithm for minimizing it.
Applying the SLA algorithm to our $BB\gamma$ loss yielded slightly superior results to the sigmoidal loss when the empirical counts from the training set for $\alpha$, $\beta$ and $n$ are used and slightly worse results when we used $\alpha=1$, $\beta=1$ and $n=100$. 
%

Analyzing the ability of different loss formulations and algorithms to minimize the 0-1 loss on different datasets using a common model class (i.e. linear models) can reveal differences in optimization performance across different models and algorithms. 
However, we are certainly more interested in evaluating the ability of different loss functions and optimization techniques to learn models that can be generalized to new data. We therefore provide the next set of experiments using traditional training, validation and testing splits, again following the protocols used in \cite{nguyen2013algorithms}; however, as we shall soon see, these experiments underscored the importance of extending the original SLA algorithm to automate the adjustment of learning parameters.

In Tables \ref{tab:sum_01_10} and \ref{tab:err_01_10}, we create 10 random splits of the data and perform a traditional 5 fold evaluation using cross validation within each training set to tune hyper-parameters. In Table \ref{tab:sum_01_10}, we present the sum of the 0-1 loss over each of the 10 splits as well as the total 0-1 loss across all experiments for each algorithm. This analysis allows us to make some intuitive comparisons with the results in Table \ref{tab:SLA_losses}, which represents an empirically derived lower bound on the 0-1 loss. In Table \ref{tab:err_01_10}, we present the traditional mean accuracy 
across these same experiments. 
Examining columns SLA vs. BBLR$^2$ in Table \ref{tab:sum_01_10}, we see that our re-formulated logistic loss is able to outperform the sigmoidal loss, but that only with the addition of the additional tuning of parameters during the optimization in column BBLR$^3$ are we able to improve upon the overall zero-one loss yielded by the logistic regression and SVM baseline methods. However, it is important to remember that all of these methods are based on an underlying \emph{linear} model, these are comparatively small datasets consisting of relatively low dimensional input feature vectors. As such, we do not necessarily expect there to be any statistically significant differences test set performance due to zero-one loss minimization performance. The same observation was made in \cite{nguyen2013algorithms} and it motivated their own exploration of learning with noisy feature vectors. We follow a similar path below, but then go on further to explore datasets that are much larger and of much higher dimensions in our subsequent experimental work.

\begin{table}[htb!]
\begin{center}
\caption{\BRkh{The 0-1 loss for all data in the datasets given on the left}{The total 0-1 loss for all data in a dataset}.  (left to right) Results using logistic regression, a linear SVM, our method with $\alpha=\beta=1$ and $n=100$ (BBLR$^1$), the sigmoid loss with the SLA algorithm and our approach with empirical values for $\alpha$, $\beta$ and $n$ (BBLR$^2$).  }
\label{tab:01_all}
\begin{tabular}{|c||c|c||c|c|c|}
\hline
     \Rkh{Dataset} & LR & SVM & BBLR$^1$ & SLA & BBLR$^2$  \\ \hline \hline
Breast & 21 & 19 & {\bf 11}    & 14 & 12  \\ \hline
Heart & 39 & 40 & 42     & 39 & {\bf 26}   \\ \hline
Liver & 102 & 100 & 102  & {\bf 90} & {\bf 90}   \\ \hline
Pima & 167 & 167 & 169   & {\bf 157} & 166  \\ \hline \hline
\BRkh{Total $L_{01}$}{Sum} & 329 & 326 & 324  & 300 & {\bf 294}  \\ \hline
\end{tabular}
\end{center}
\end{table}

\begin{table}[htb!]
\begin{center}
\caption{The sum of the mean 0-1 loss over 10 repetitions of a 5 fold leave one out experiment. (left to right) Performance using logistic regression, a linear SVM, the sigmoid loss with the SLA algorithm, our BBLR model with optimization using the SLA optimization algorithm and our BBLR models (BBLR$^2$ and BBLR$^3$) with additional tuning of the modified SLA algorithm.  }
\label{tab:sum_01_10}
\begin{tabular}{|c||c|c||c|c|c|}
\hline
                & LR & SVM & SLA & BBLR$^2$ & BBLR$^3$  \\ \hline \hline
Breast          & 22 & {\bf 21} & 23     & 22 & {\bf 21}  \\ \hline
Heart           & 45  & 45 & 48    & 50 & {\bf 43}   \\ \hline
Liver           & 109 & 110 & 114  & {\bf 105} & {\bf 105}   \\ \hline
Pima            & 172  & 172 & 184 & 176 & {\bf 171}  \\ \hline \hline
Total $L_{01}$  & 348 & 348 & 368  & 354 & {\bf 340}  \\ \hline
\end{tabular}
\end{center}
\end{table}

\begin{table}[htb!]
\begin{center}
\caption{The errors ($\%$) averaged across the 10 test splits of a 5 fold leave one out experiment. (left to right) Performance using logistic regression, a linear SVM, the sigmoid loss with the SLA algorithm, our BBLR model with optimization using the SLA optimization algorithm and our BBLR model with additional tuning of the modified SLA algorithm.  }
\label{tab:err_01_10}
\begin{tabular}{|c||c|c||c|c|c|c|}
\hline
                & LR & SVM & SLA & BBLR$^2$ & BBLR$^3$ & BBLR$^4$  \\ \hline \hline
Breast          & 3.2 &  3.1    & 3.6     & 3.2 & 3.1 &  {\bf3.0}\\ \hline
Heart           & 16.8  & 16.6 & 17.7    & 18.6 & 15.9 & {\bf 15.7} \\ \hline
Liver           & 31.5  & 31.8 & 32.9   & 30.6 & {\bf 30.4} & 30.5  \\ \hline
Pima            & 22.3  & 22.4  & 23.9  & 23.0 & {\bf 22.2} & {\bf 22.2}\\ \hline \hline
\end{tabular}
\end{center}
\end{table}

\begin{table}[htb!]
\begin{center}
\caption{The sum of the mean 0-1 loss over 10 repetitions of a 5 fold leave one out experiment where 10\% noise has been added to the data following the protocol given in \cite{nguyen2013algorithms}. (left to right) Performance using logistic regression, a linear SVM, the sigmoid loss with the SLA algorithm, our BBLR model with optimization using the SLA optimization algorithm and our BBLR model with additional tuning of the modified SLA algorithm. We give the relative improvement in error of the BBLR$^3$ technique over the SVM in the far right column. }
\label{tab:sum_mean_loss01}
\begin{tabular}{|c||c|c||c|c|c||c|}
\hline
                & LR & SVM & SLA & BBLR$^2$ & BBLR$^3$ & Impr.  \\ \hline \hline
Breast          & 36 & 34 & 26     & 26 & {\bf 25} & 26\% \\ \hline
Heart           & 44  & 44 & 49    & 47 & {\bf 42} & 4\%  \\ \hline
Liver           & 150 & 149 & 149  & 149 & {\bf 117} & 21\%   \\ \hline
Pima            & 192  & 199 & 239 & 185 & {\bf 174} & 12\% \\ \hline \hline
Total $L_{01}$  & 422 & 425 & 463  & 374 & {\bf 359} & 16\% \\ \hline
\end{tabular}
\end{center}
\end{table}

\begin{table}[htb!]
\begin{center}
\caption{The errors ($\%$) averaged over 10 repetitions of a 5 fold leave one out experiment in which 10\% noise has been added to the data. (left to right) Performance using logistic regression, a linear SVM, the sigmoid loss with the SLA algorithm, our BBLR model with optimization using the SLA optimization algorithm and our BBLR model with additional tuning of the modified SLA algorithm.  }
\label{tab:avg_err_rates}
\begin{tabular}{|c||c|c||c|c|c|c|}
\hline
                & LR & SVM & SLA & BBLR$^2$ & BBLR$^3$ & BBLR$^4$ \\ \hline \hline
Breast          & 5.2 & 5.0    & 3.8     & 3.9 & 3.7 & {\bf 3.4} \\ \hline
Heart           & 16.4  & 16.2 & 18.1    & 17.3 & 15.5 & {\bf 15.2}  \\ \hline
Liver           & 43.5  & 43.1 & 43.3   & 33.8 & 34.1 & {\bf 34.0}  \\ \hline
Pima    & 25.0  & 25.9  & 31.1  & 24.0 & 22.7 & {\bf 22.5} \\ \hline \hline
\end{tabular}
\end{center}
\end{table}

In Table \ref{tab:sum_mean_loss01}, we present the sum of the mean 0-1 loss over 10 repetitions of a 5 fold leave one out experiment where 10\% noise has been added to the data following the protocol given in \cite{nguyen2013algorithms}. Here again, our BBLR$^2$ achieved a moderate gain over the SLA algorithm, whereas the gain of BBLR$^3$ over other models is noticeable. In this table, we also show the percentage of improvement for our best model over the linear SVM. In Table \ref{tab:avg_err_rates}, we show the average errors ($\%$) for these 10\% noise added experiments. We see here that the advantages of more directly approximating the zero one loss are more pronounced. However, the fact that the SLA approach failed to outperform the LR and SVM baselines in our experiments here; whereas in a similar experiment in \cite{nguyen2013algorithms} the SLA algorithm and sigmoidal loss did outperform these methods leads us to believe that the issue of per-dataset learning algorithm parameter tuning is a significant issue. However, we observe that our BBLR$^2$ experiment which used the original SLA optimization algorithm outperformed the sigmoidal loss function optimized using the SLA algorithm. These results support the notion that our proposed Beta-Bernoulli logistic loss is in itself a superior approach to approximate the zero-one loss from an empirical perspective. However, our results in column  BBLR$^4$ indicate that the combined use of our novel logistic loss and learning parameter optimization yield the most substantial improvements to zero-one loss minimization, or correspondingly improvements to accuracy.
\subsubsection{Pooled McNemar Tests :}
We performed McNemar tests for the four UCI benchmarks comparing BBLR$^3$ with LR and linear SVMs. As we do not have significant number of test instances for any of these benchmarks, it became difficult to statistically justify and compare results. Therefore, we performed pooled McNemar tests by considering each split of our 5-fold leave one out experiments as independent tests and collectively performing the significance tests as a whole. The results of this pooled McNemar test is given in Table \ref{tab:pooled_mcnemar}. Interestingly, for our noisy dataset experiments, our BBLR$^3$ was found to be statistically significant over both the LR and SVM models with $p\le 0.01$.     
\begin{table}[htb!]
\begin{center}
\caption{z-static for pooled McNemar tests; values $\ge 2.32$ denotes statistically significance with $p \le 0.01$.}
\label{tab:pooled_mcnemar}
\begin{tabular}{|c||c|c|}
\hline
& BBLR$^3$ vs. LR & BBLR$^3$ vs. SVM\\
\hline
clean-UCI & 3.17 & 0.69 \\
\hline
noisy-UCI & 4.33 & 3.7\\
\hline
\hline
\end{tabular}
\end{center}
\end{table}

\subsubsection{Experiments with LibSVM Benchmarks}

In this section, we present classification results using two 
much larger datasets: the web8, and the webspam-unigrams. These datasets have predefined  training and testing splits, which are distributed on the web site accompanying \cite{zhang2011smoothing}\footnote{http://users.cecs.anu.edu.au/~xzhang/data/}. 
These benchmarks are also distributed through the LibSVM binary data collection\footnote{http://www.csie.ntu.edu.tw/˜cjlin/libsvmtools/datasets/binary.html}. The webspam unigrams data originally came from the study in \cite{wang2012evolutionary}\footnote{http://www.cc.gatech.edu/projects/doi/WebbSpamCorpus.html}.
\Rkh{Table \ref{tab:libsvm_large_scale} compiles some details of thsese databases.}
%

\begin{table}[htb!]
\begin{center}
\caption{Standard larger scale LibSVM benchmarks used for our experiments; $n_+:n_-$ denotes the ratio of positive and negative training data.}
\label{tab:libsvm_large_scale}
\begin{tabular}{|c||c|c|c|c|}
\hline
Dataset & \# Examples & \# Dim. & Sparsity (\%) & $n_+:n_-$ \\ \hline \hline 
web8 & 59,245 & 300 & 4.24 & 0.03 \\ \hline
webspam-uni & 350,000 & 254 & 33.8 & 1.54\\ \hline
\end{tabular}
\end{center}
\end{table}

For these experiments we do \emph{not} add additional noise to the feature vectors.
In Table \ref{tab:large_scale}, we present classification results, and one can see that 
%
for both cases our BBLR$^3$ approach shows improved performance over the LR and the linear SVM baselines. As in our earlier small scale experiments, we used our own LR implementation and the liblinear SVM for these large scale experiments.

\begin{table}[htb!]
\begin{center}
\caption{Errors ($\%$) for larger scale experiments on the data sets from the LibSVM evaluation archive. When BBLR$^3$ is compared to a model using McNemer's test, $^{**}$ : BBLR$^3$ is statistically significant with a $p$ value $\le 0.01$}
\label{tab:large_scale}
\begin{tabular}{|c||c|c|c|}
\hline
Data set & LR & SVM & BBLR$^3$ \\\hline \hline 
web8 & 1.11$^{**}$ & 1.13$^{**}$ & {\bf 0.98} \\ \hline
webspam-unigrams & 7.26$^{**}$ & 7.42$^{**}$ & ${\bf 6.56}$ \\ \hline
\end{tabular}
\end{center}
\end{table}

\Rkh{
We performed McNemar's statistical tests comparing our BBLR$^3$ with LR and linear SVM models for these two datasets. The results are found to be statistically significant with a $p$ value $\le$ 0.01 for all cases. Given that no noise has been added to these widely used benchmark problems and that each method compared here is fundamentally based on a linear model, the fact that these experiments show statistically significant improvements for BBLR$^3$ over these two widely used methods is quite interesting.
}

\subsubsection{Experiments with Product Reviews}

The goal of these tasks are to predict whether a product review is either positive or negative. 
For this set of experiments, we used the count based unigram features for four databases from the website associated with \cite{dredze2008confidence}. 
%
%
For each database, there are 1,000 positive and 1,000 negative product reviews.
\Rkh{Table \ref{tab:bbbb_details} compiles the feature dimension size of these sparse databases.}
\Rkh{
\begin{table}[htb!]
\begin{center}
\caption{Standard product review benchmarks  used in our experiments.}
\label{tab:bbbb_details}
\begin{tabular}{|c|c|c|}
\hline
Dataset & Database size & Feature dimensions \\ \hline \hline 
Books &  & 28,234 \\
DVDs  & 2000 & 28,310 \\
Electronics & & 14,943 \\
Kitchen & & 12,130 \\
\hline
\hline
\end{tabular}
\end{center}
\end{table}
}

We present results in Table \ref{tab:sum_mean_loss_bbbb} using a ten fold cross validation setup as performed by \cite{dredze2008confidence}. Here again we do \emph{not} add noise the the data.


\begin{table}[htb!]
\begin{center}
\caption{Errors ($\%$) on the test sets. When BBLR$^3$ and BBLR$^4$ are compared to LR and an SVM using McNemer's test, $^{*}$ the results are statistically significant with a $p$ value $\le 0.05$.}
\label{tab:sum_mean_loss_bbbb}
\begin{tabular}{|c||c|c|c|c|}
\hline
                & Books & DVDs & Electronics & Kitchen   \\ \hline \hline
LR          & 19.75 & 18.05$^*$ & 16.4      & 13.5  \\ \hline
SVM           & 20.45 & 21.4$^*$ & 17.75    & 14.6   \\ \hline
BBLR$^3$           &  18.38 &  17.5 & 16.29  & {\bf 13.0}    \\ \hline
BBLR$^4$           & {\bf 18.15} & {\bf 16.8} & {\bf 15.21}  & {\bf 13.0}    \\\hline
\end{tabular}
\end{center}
%
\end{table}


For all four databases, our BBLR$^3$ and BBLR$^4$ models outperformed both the LR and linear SVM. To further analyze these results, we also performed a McNemer's test. 
%
%
For the Books and the DVDs database,  the results of our BBLR$^3$ and BBLR$^4$ models are found statistically significant over both the LR and linear SVM with a $p$ value $\le 0.05$. 
BBLR$^4$ tended to outperform BBLR$^3$, but not in a statistically significant way. However, since the primary advantage of the BBLR$^4$ configuration is that it yields more accurate probabilities, we to not necessarily expect it to have dramatically superior performance compared to BBLR$^3$ for classification. For this reason we explore the problem of using such models in the context of a structured prediction in the next set of experiments. When BBLR models are used to make structured predictions our hypothesis is that the benefits of providing a more accurate probabilistic prediction should be apparent through improved joint inference.


\subsection{Structured Prediction Experiments}

One of the advantages of our Beta-Bernoulli logistic loss is that it allows a model to produce more accurate probabilistic estimates. Intuitively, the controllable nature of the plateaus in the log probability view of our formulation allow probabilistic predictions to take on values that are more representative of an appropriate confidence level for a classification. In simple terms, predictions for feature vectors far from a decision boundary need not take on values that are near probablity zero or probability one when the Beta-Bernoulli logistic model is used. If such models are used as components to larger systems which uses probabilistic inference for more complex reasoning tasks, the additional flexibility could be a significant advantage over the traditional logistic function formulation. The following experiments explore this hypothesis.

\begin{figure*}[htb!]
\begin{center}
%
%
\begin{tabular}{p{7.5cm} p{7.5cm} }
%
\hline
\hspace{3.3cm}   \hspace{0.1cm} Image 1 & \hspace{2.1cm} Image 2\\
\hspace{2.01cm}   \hspace{.90cm} \includegraphics[scale=0.15]{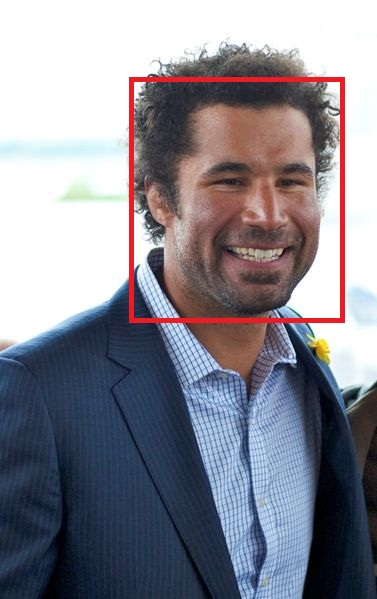} & \includegraphics[scale=0.20]{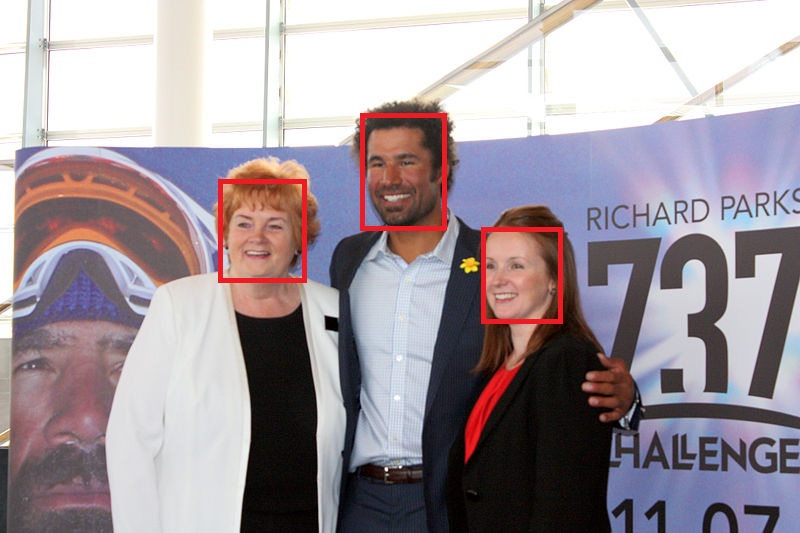}\\

\hspace{0.01cm} Image source: \hspace{0.01cm} Info-box & Body text \\[0.2cm]
\hspace{0.01cm} File name : \hspace{0.55cm} \textbf{Richard\_Parks}.jpg & 737 Challenge.jpg\\[0.2cm]
\hspace{0.01cm} Caption text : \hspace{0.08cm} NULL & \textbf{Richard Parks} celebrating the end of the 737 Challenge at the National Assembly for Wales on 19 July 2011\\[0.2cm]
%
\hline
\vtop{\null{\hbox{\includegraphics[scale=0.26]{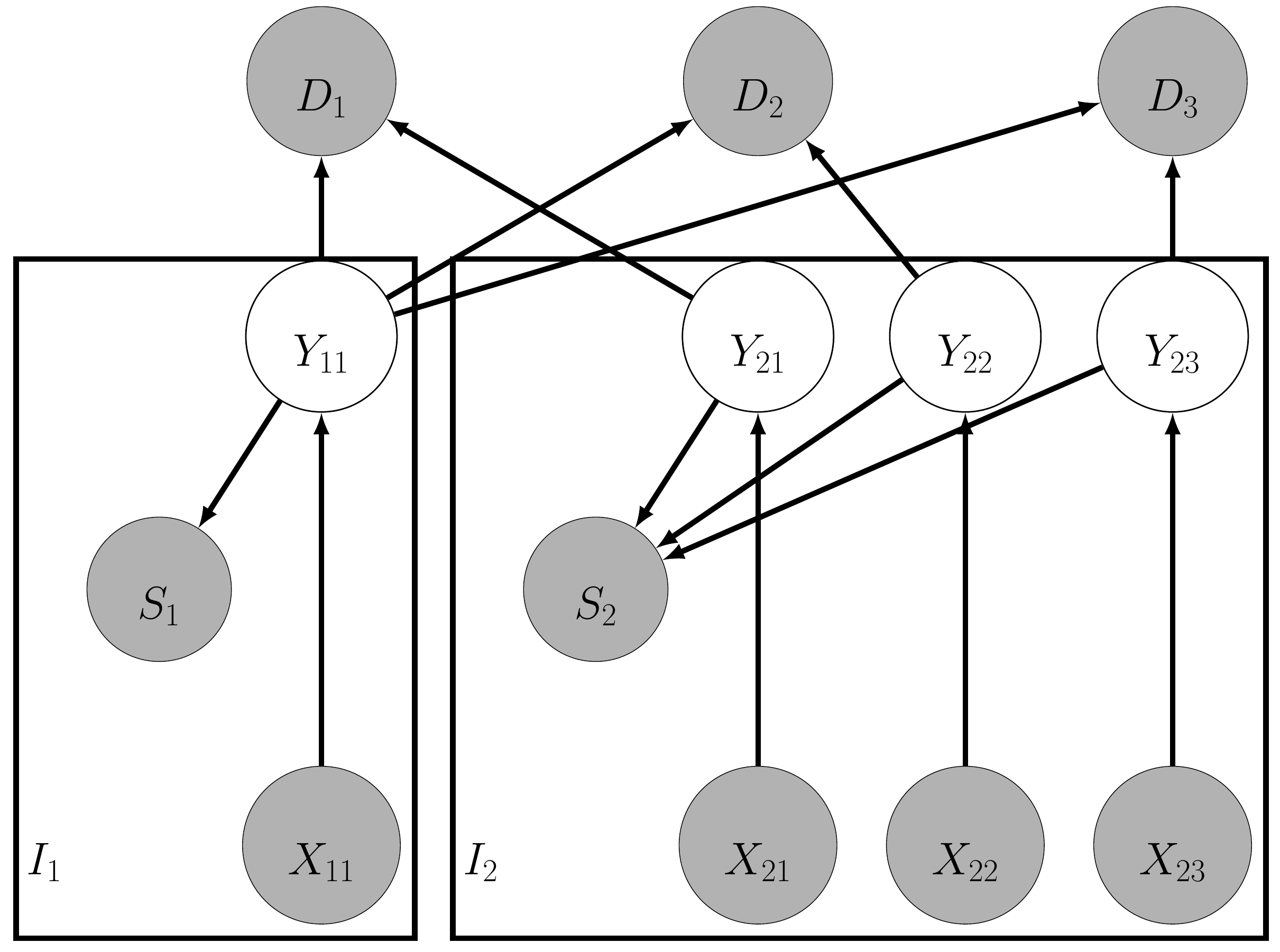}}}} & \vtop{\null{\hbox{\begin{tabular}{p{1.5cm} p{4cm}}
Variables & Description\\[0.3cm]
\hline
$D_{l}$ :& Visual similarity for a pair of faces, $x_{mn}$ and $x_{m^\prime n^\prime}$, across different images.\\[0.2cm]
$Y_{mn} :$ & Binary target vs. not target label for face, $x_{mn}$. \\ [0.2cm]
$S_m$ : & Constraint variable for image $m$. \\[0.2cm]
$X_{mn} :$ & Local features for a face. \\
%
\end{tabular}}}}\\
%
\hline
\end{tabular}
\caption{(First row) : Images, face detection results through bounding boxes, and corresponding text and meta information from the Wikipedia biography page for Richard Parks. (Bottom row) : An instance of our facial co-reference model and its variable descriptions.
}
\label{fig:mining_model_gen}
\end{center}
\end{figure*}

In \cite{AAAI14}, we performed a set of face mining experiments from Wikipedia biography pages using a technique that relies on probabilistic inference in a joint probability model. 
\CRkh{
%
For a given identity, our mining technique dynamically creates probabilistic models to disambiguate the faces that correspond to the identity of interest. These models integrate uncertain information extracted throughout a document arising from three different modalities: text, meta data and images. Information from text and metadata is integrated into the larger model using multiple logistic regression based components. 

The images, face detection results as bounding boxes, some text and meta information extracted from one of the Wikipedia identity, Mr. Richard Parks, are shown in the top panel of Figure \ref{fig:mining_model_gen}. In the bottom panel, we show an instance of our mining model and give a summary of the variables used in our technique. The model is a dynamically instantiated Bayesian network. Using the Bayesian network illustrated in Figure \ref{fig:mining_model_gen}, the processing of information is intuitive. Text and meta-data features are taken as input to the bottom layer of random variables $\{X\}$, which influence binary (target or not target) indicator variables $\{Y\}$ for each detected face through logistic regression based sub-components. The result of visual comparisons between all faces detected in different images are encoded in the variables $\{D\}$.
}
%

\CRkh{
   Both text and meta data are transformed into feature vectors associated with each detected instance of a face. 
For text analysis, we use information such as: image file names and image captions.  The location of an image in the page is an example of what we refer to as meta-data. We also treat other information about the image that is not directly involved in facial comparisons as meta-data, ex. the relative size of a face to other faces detected in an image.  The bottom layer or set of random variables $\{X\}$ in Figure \ref{fig:mining_model_gen} are used to encode these features, and we discuss the precise nature and definition of these features in more detail in  \cite{AAAI14}. $X_{mn}= [X_1^{(mn)},X_2^{(mn)},\cdots X_K^{(mn)}]^T$ is therefore the local feature vector for a face, $x_{mn}$, where $X_k^{(mn)}$ is the  $k^{th}$ feature for face index $n$ for image index $m$. 
These features are used as the input to the part of our model responsible for producing the probability that a given instance of a face belongs to the identity of interest, encoded by the random variables $\{Y\}$ in Figure \ref{fig:mining_model_gen}. $\{Y\}=\{\{Y_{mn}\}_{n=1}^{N_m}\}_{m=1}^M$ is therefore a set of binary target vs. not target indicator variables corresponding to each face, $x_{mn}$. Inferring these variables jointly corresponds to the goal of our mining model.
}
%
%
The joint conditional distribution defined by the general case of our model is given by

\begin{align}
p(\{\{Y_{mn}&\}_{n=1}^{N_m}\}_{m=1}^M,\{D_l\}_{l=1}^L,\{S_m\}_{m=1}^M|\{\{X_{mn}\}_{n=1}^{N_m}\}_{m=1}^M)\nonumber \\
= &\prod_{m=1}^{M}\prod_{n=1}^{N_m} p(Y_{mn}| X_{mn}) p(S_{m}|\{Y_{mn^\prime}\}_{n^\prime=1}^{N_m^\prime}) \nonumber \\
&\prod_{l=1}^Lp(D_{l}|\{Y_{{m^\prime_l}{n^\prime_l}},Y_{{m^{\prime\prime}_l}{n^{\prime\prime}_l}}\}). 
%
\label{eqn:single_model_eq1}
\end{align}

Apart from comparing cross images faces, $p(D_{l}|\{Y_{{m^\prime_l}{n^\prime_l}},Y_{{m^{\prime\prime}_l}{n^{\prime\prime}_l}}\})$, the joint model uses predictive scores from per face local binary classifiers, $p(Y_{mn}| X_{mn})$. As mentioned above and discussed in more detail in \cite{AAAI14}, we used Maximum Entropy Models (MEMs) or Logistic Regression models for these local binary predictions working on multimedia features in our previous work.
%

%
%
%
Here, we compare the result of replacing the logistic regression components in the model discussed above with our BBLR formulation. We examine the impact of this change in terms of making predictions based solely on independent models taking text and meta-data features as input as well as the impact of this difference when LR vs BBLR models are used as sub-components in the joint structured prediction model. 
Our hypothesis here is that the BBLR method might improve results due to its robustness to outliers (which we have already seen in our binary classification experiments) and that the method is potentially able make more accurate probabilistic predictions, which could in turn lead to more precise joint inference.

\cut{
\begin{table}[htb!]
\begin{center}
\begin{tabular}{|c|c|c|}
\hline
& Text-only features & Joint model with aligned faces \\
\hline
MEM & 64.9  & 76.21  \\
\hline
BBLR & 67.25  & \textbf{78.03} \\
\hline
\end{tabular}
\end{center}
\caption{Comparing MEM and BBLR when used in structured prediction problems. Showing their accuracies (\%) and standard Deviation  }
\label{tab:bblr_structured}
\end{table}
} 

\begin{table}[htb!]
\begin{center}
\begin{tabular}{|c|c|c|}
\hline
& Text-only features & Joint model with aligned faces \\
\hline
MEM & 63.4  & 76.0  \\
\hline
BBLR$^3$ & 67.8  & 78.2 \\
\hline
BBLR$^4$ & 70.2$^*$  & \textbf{81.5}$^*$ \\
\hline
\end{tabular}
\end{center}
\caption{Comparing MEM and BBLRs when used in structured prediction problems. Showing their accuracies (\%)} 
\label{tab:bblr_structured}
\end{table}

For this particular experiment, we use the biographies with 2-7 faces. 
Table \ref{tab:bblr_structured} shows results comparing the MaxEnt model with our BBLR model. 
The results are for a five-fold leave one out of the wikipedia dataset. 
One can see that we do indeed obtain superior performance with the independent BBLR models over the Maximum Entropy models. We also see improvement to performance when BBLR models are used in the coupled model where joint inference is used for predictions. 

In the row labelled BBLR$^4$, we optimized $\{ w,\theta_B,\gamma, \lambda\}$ in addition to other model parameters using the technique, explained in Section \ref{sec:w_ThB}. This produced statistically significant results compared to the maximum entropy model with $p\le 0.05$. For this significance test, we used the McNemar test like our earlier sets of experiments.

\subsection{Kernel Logistic Regression with the Generalized Beta-Bernoulli Loss}

In Table \ref{tab:kernel_compare_svs} we compare Beta-Bernoulli logistic regression with an SVM and Kernel Beta-Bernoulli logistic regression (KBBLR). We see that our proposed approach compare favorably to the SVM result which is widely considered as a state of the art, strong baseline.

\begin{table}[htb!]
\begin{center}
\caption{Comparing Kernel BBLR with an SVM and linear BBLR on the standard UCI datasets (no sparsity).}
\label{tab:kernel_compare_svs}
\begin{tabular}{|c||c|c|c|c|}
\hline
Dataset & BBLR & SVM & KBBLR \\\hline \hline 
Breast & ${\bf 2.82}$ & $3.26$ & $2.98$ \\ \hline
Heart & $17.08$ & $17.76$ & ${\bf 16.27}$ \\ \hline
Liver & $31.80$ & $29.61$ & ${\bf 26.91}$ \\ \hline
Pima & ${\bf 21.57}$ & $22.44$ & $22.9$ \\ \hline
\end{tabular}
\end{center}
\end{table}

\subsection{Sparse Kernel BBLR} 


As shown in \cite{collobert2006trading}, one of the advantages of using the ramp loss for kernel based classification is that it can yield models that are even sparser than traditional SVMs based on the hinge loss. It is well known that $L_2$ based regularization does not typically yield sparse solutions when used with traditional kernel logistic regression.
Our analysis of the previous experiments reveals that the $L_2$ regularized smooth zero one loss approximation approach proposed here does not in general lead to sparse models as well. The well known $L_1$ or lasso regularization method can yield sparse solutions, but often at the cost of prediction performance. Recently the so called elastic net regularization approach \cite{zou2005regularization} based on a weighted combination of $L_1$ and $L_2$ regularization has been shown more effective at encouraging sparsity with a less negative impact on performance. The elastic net approach of course can be viewed as a prior consisting of the product of a Gaussian and a Laplacian distribution. However, part of the motivation for the use of these methods is that they yield convex optimization problems when combined with the log logistic loss. Since we have developed a robust approach for optimizing a non-convex objective function above, this opens the door to the use of non-convex sparsity encouraging regularizers. 
Correspondingly, we propose and explore below a prior on parameters, or equivalently, a novel regularization approach based on a mixture of a Gaussian and a Laplacian distribution. This formulation can behave like a smooth approximation to an $L_0$ counting ``norm'' prior on parameters in the limit as the Laplacian scale parameter goes to zero and the Gaussian variance goes to infinity. 

With a (marginalized) Gaussian-Laplace mixture prior, our KBBLR log-likelihood becomes
\begin{align}
\label{eqn:gauss_l_mixtureX}
\cal{L}(D|{\bf a}) &=\sum_{i=1}^n \big( {y_i} \log \mu_{i} + (1-{y_i}) \log (1 - \mu_{i}) \big) \\\nonumber
&+\sum_j\ln \big(\pi_g \mathcal{N}(a_j;0,\sigma_g^2)+\pi_l\mathcal{L}(a_j;0,b_l)\big)
\end{align}
where $\mu_i=\mu_{K\beta B}({\bf a},{\bf x}_i)$ is our kernel Beta-Bernoulli model as defined in section \ref{sec:bb_klr}, equation (\ref{eqn:KBBLR_mu}). 
For each $a_j$, its prior is modeled through a mixture of a zero mean Gaussian $\mathcal{N}(a_j;0,\sigma_g^2)$ with variance $\sigma_g^2$ and a Laplacian distribution $\mathcal{L}(a_j;0,b_l)$, located a zero with shape parameter $b_l$. For convenience we give the relevant partial derivatives for this prior in Appendix \ref{kbblr:sparse}. In our approach we also optimize the hyper-parameters $\{\pi_g,\pi_l,\sigma_g^2,b_l\}$ of this prior using hard assignment Expectation Maximization steps that are performed after 
step 3 of Algorithm 2. For precision we outline the steps of the modified range-optimization for Kernel BBLR (KBBLR) in Algorithm 3 found in  Appendix C.

In Table \ref{tab:kernel_compare}, we compare 
sparse KBBLR and the SVM using a Radial Basis Function (RBF) kernel. 
The SVM free parameters were tuned by a cross validation run over the training data. 
For a sparse KBBLR solution, we used a mixture of a Gaussian and a Laplacian prior on the kernel weight parameters as presented above.

\begin{table}[htb!]
\begin{center}
\caption{Comparing sparse kernel BBLR with SVMs on the standard UCI evaluation datasets.}
\label{tab:kernel_compare}
\begin{tabular}{|c||p{1.0cm}|p{1.7cm}|p{1.40cm}|p{1.9cm}|}
\hline
Dataset & SVM & Avg. Support Vectors & Sparse KBBLR & Avg. Support Vectors \\\hline \hline 
Breast & $3.26$ & {\bf 107} & ${\bf 2.83}$ & 127 \\ \hline
Heart & $17.76$ & 148 & ${\bf 16.40}$ & {\bf 85}\\ \hline
Liver & $29.61$ & 269 & ${\bf 28.74}$ & {\bf 111}\\ \hline
Pima & ${\bf 22.44}$ & 548 & $23.52$ &  {\bf 269} \\ \hline
\end{tabular}
\end{center}
\end{table}

\begin{figure}[ht!]
\centering
\includegraphics[scale=0.17]{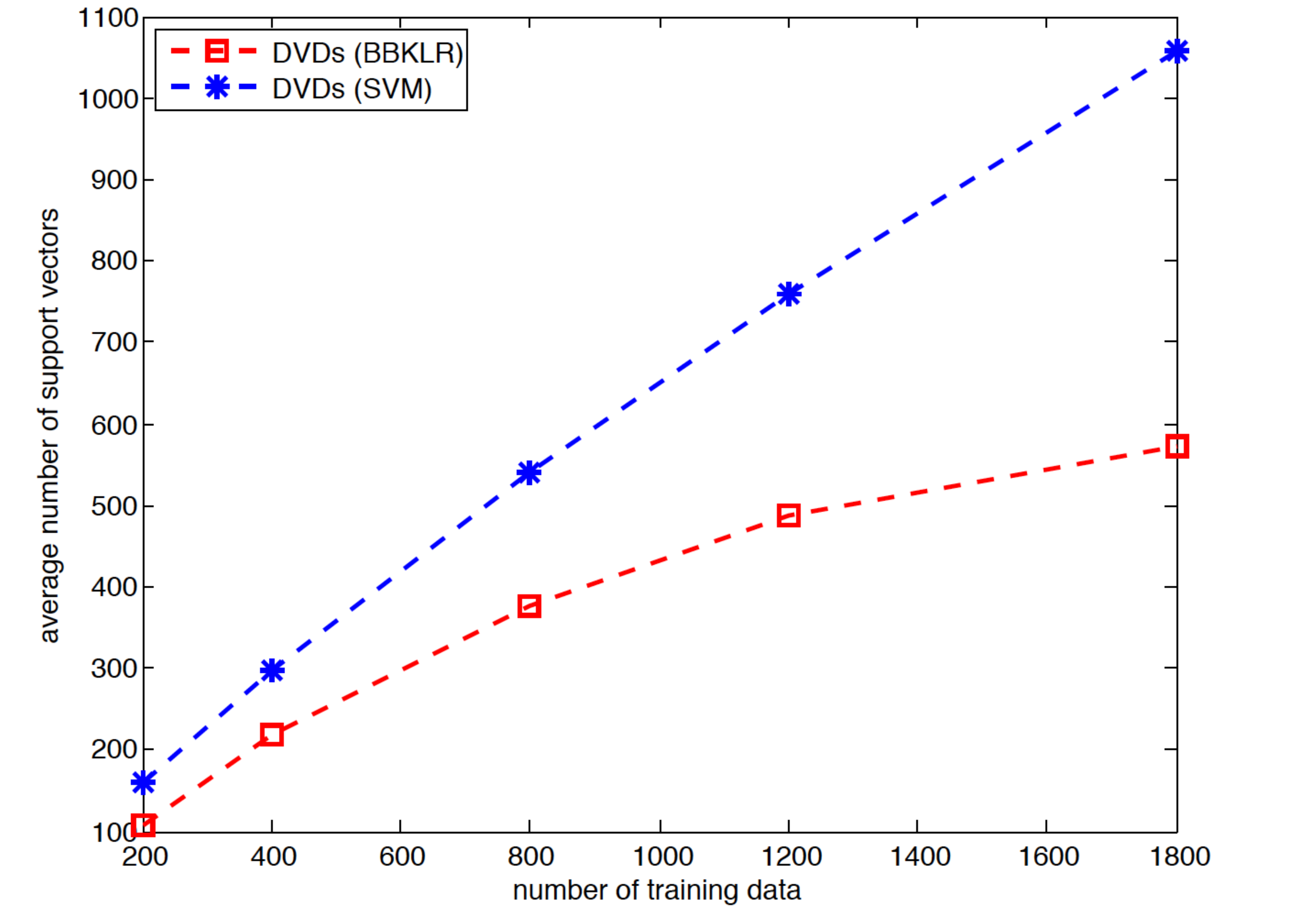}
\includegraphics[scale=.17]{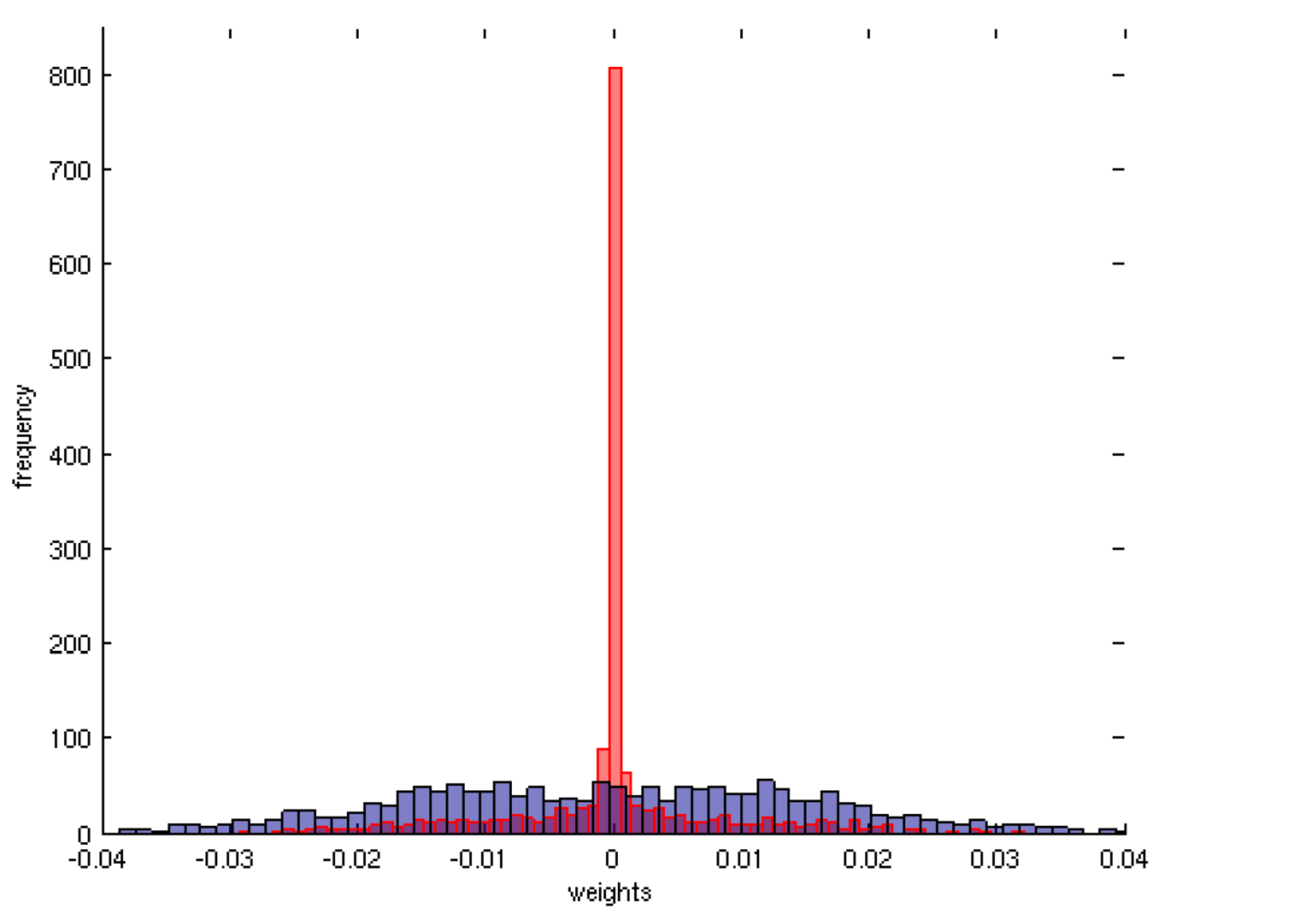}
\caption{(left image) Comparing sparse KBBLR with SVMs : the vertical axis shows the increase in the number of support vectors for an increase in the number of training instances (horizontal axis). (right image) This image shows the weight distribution for the L2 regularized KBBLR (in blue) and the final Gauss-Laplacian mixture solution (in red) at the end of optimization.}
\label{fig:kbblr_vs_svms}
\end{figure}


Table \ref{tab:kernel_compare} compares sparse Kernel BBLR with SVMs on the standard UCI datasets. Figure \ref{fig:kbblr_vs_svms} shows trends in the sparsity curves for an increase in the number of training instances comparing KBBLR with SVMs for one of the product review databases. We can see that KBBLR scales up well compared to an SVM solution when training data size increases. Support vectors for SVMs increase almost linearly for an increase in the database size, an effect that has been confirmed in a number of other studies \cite{steinwart2003sparseness,collobert2006trading}. In comparison we can see that KBBLR with a Gaussian-Laplacian mixture prior produces a logarithmic curve for an increase in the database size. The right panel of the same figure also shows the weight distribution before and after the KBBLR optimization with a Gaussian-Laplacian mixture prior which yields the observed sparse solution.

\section{Discussion and Conclusions}
\label{sec:Discuss_Conc}
We have presented a novel formulation for learning with an approximation to the zero one loss. Through our generalized Beta-Bernoulli formulation, we have provided both a new smooth 0-1 loss approximation method and a new class of probabilistic classifiers. 
%
Our experimental results indicate that our generalized Beta-Bernoulli formulation is capable of yielding superior performance to traditional logistic regression and maximum margin linear SVMs for binary classification. 
Like other ramp like loss functions one of the principal advantages of our approach is that it is more robust dealing with outliers compared to traditional convex loss functions. 
%
%
%
Our modified SLA algorithm, which adds a learning hyper-parameter optimization step shows improved performance over the original SLA optimization algorithm in \cite{nguyen2013algorithms}. 

We have also presented and explored a kernelized version of our approach which yields performance competitive with non-linear SVMs for binary classification. 
Furthermore, with a Gaussian-Laplacian mixture prior on parameters our kernel Beta-Bernoulli model is able to yield sparser solutions than SVMs while retaining competitive classification performance. Interestingly, for an increase in training database size, our approach exhibited logarithmic scaling properties which compares favourably to the linear scaling properties of SVMs. To the best of our knowledge this is the first exploration of a Gauss-Laplace mixture prior for parameters -- certainly in combination with our novel smooth zero-one loss formulation. The ability of this prior to behave like a smooth approximation to a counting prior is similar to an approach known as bridge regression in statistics. However, our mixture formulation has more flexibility compared to the simpler functional form of bridge regression. Interestingly, the combination of our generalized Beta-Bernoulli loss with a Gaussian-Laplacian parameter prior can be though of a smooth relaxation to learning with a zero one loss and an $L_0$ counting prior or regularization -- a formulation for classification that is intuitively attractive, but has remained elusive in practice until now. 

\Rkh{We also tested our generalized Beta-Bernoulli models for a structured prediction task arising from the problem of face mining in Wikipedia biographies. Here also our model showed better performance than traditional logistic regression based approaches, both when they were tested as independent models, and when they were compared as sub-parts of a Bayesian network based structured prediction framework. This experiment shows signs that the model and optimization approach proposed here may have further potential to be used in complex structured prediction tasks. 
}

\section*{Acknowledgements}
We thank the NSERC Discovery Grants program and Google for a Faculty Research Award which helped support this work.

\bibliographystyle{abbrv}
\bibliography{smooth_loss}

\appendix

\section{Experimental Details}
\label{appendix:sla}

In the interests of reproducibility, we also list below the algorithm parameters and the recommended settings as given in \cite{nguyen2013algorithms} :

%
\begin{description}
\itemsep0pt \parskip1pt \parsep1pt
  \item $r_{R}=2^{-1}$, a search radius reduction factor;
  \item $R_0=8$, the initial search radius;
  \item $r_\epsilon=2^{-1}$, a grid spacing reduction factor;
  \item $\epsilon_{S_0}=0.2$, the initial grid spacing for 1-D search;
  \item $r_\gamma=10$, the gamma parameter reduction factor;
  \item $\gamma_{MIN}=2$, the starting point for the search over $\gamma$;
  \item $\gamma_{MAX}=200$, the end point for the search over $\gamma$.
\end{description}

As a part of the Range Optimization procedure there is also a standard gradient descent procedure using a slowly reduced learning rate. The procedure has the following specified and unspecified default values for the constants defined below:
\begin{description}
\itemsep0pt \parskip1pt \parsep1pt
  \item $r_{G}=0.1$, a learning rate reduction factor;
  \item $r_{G_{MAX}}$, the initial learning rate;
  \item $r_{G_{MIN}}$, the minimal learning rate; 
  \item $\epsilon_L$, used for a while loop stopping criterion based on the smallest change in the likelihood;
  \item $\epsilon_G$, used for outer stopping criterion based on magnitude of gradient
\end{description}

\section{Gradients for a Gaussian-Laplacian Mixture Prior} \label{kbblr:sparse}

\cut{
\begin{figure}[ht!]
\centering
\includegraphics[scale=.25]{he_gl.pdf}
\includegraphics[scale=.25]{gl_he0.pdf}\\
\includegraphics[scale=.25]{pr_gl.pdf}
\includegraphics[scale=.25]{gl_pr0.pdf}\\
\caption{For each row, the first image shows the weight distribution for the L2 regularized KBBLR (in blue) and the Gaussian-Laplace mixture solution (in red). The right figure shows the mixture model at the end of our optimization.}
\label{fig:l2_vs_gauss_laplace}
\end{figure}
} 

%
The gradient of the KBBLR likelihood is given in section \ref{sec:bb_klr}. Below we provide the gradient of the log Gaussian-Laplace mixture prior or regularization term, $R=\sum_j\ln (\pi_g \mathcal{N}(a_j;0,\sigma_g)+\pi_l\mathcal{L}(a_j;0,b_l))$

\begin{align}
\frac{dR}{da_i}= 
\sum_i\frac{\pi_g\frac{d}{da_i}\mathcal{N}(a_i;0,\sigma_g)+\pi_l\frac{d}{da_i}\mathcal{L}(a_i;0,b_l)}{\pi_g \mathcal{N}(a_i|0,\sigma_g)+\pi_l\mathcal{L}(a_i|0,b_l)}
\label{eqn:gauss_laplace_mixture_gd}
\end{align}

\begin{align}
\frac{d}{da_i}\mathcal{N}(a_i;0,\sigma_g)&=%
-\frac{a_i}{\sigma_g^3\sqrt{2\pi}}\exp(-\frac{a_i^2}{2\sigma_g^2})
\end{align}

\begin{align}
\frac{d}{da_i}\mathcal{L}(a_i;0,b_l)&=%
-\frac{1}{2b_l^2}\exp(-\frac{|a_i|}{b_l})\frac{d}{da_i}(|a_i|)
\end{align}

\[
  \frac{d}{da_i} (|a_i|)= \left\{
  \begin{array}{l l l}
     1 & \quad \text{if $a_i>0$} \\
     0 & \quad \text{if $a_i==0$}\\ 
    -1  & \quad \text{if $a_i<0$}\\
  \end{array} \right.
\]


\section{Algorithm Modifications for Sparse Gauss-Laplace KBBLR} \label{kbblr:sparse_alg}

\begin{algorithm}[htb!]
\caption{Range Optimization for Sparse KBBLR}
\label{Algorithm:SLA_range_kbblr}
\begin{algorithmic}[1]
\Require ${\bf w}, {\bf X}, {\bf t}, \gamma$, radius $R$, step size $\epsilon_S$
\Ensure Updated estimate for ${\bf w}^*$,  minimizing 0-1 loss.
\Function{Grad-Desc-In-Range-KBBLR}{${\bf w}, {\bf X}, {\bf t}, \gamma$, $R$, $\epsilon_S$}
\Repeat 
\Statex \hspace{1cm}$\rhd$\textmd{ Stage 1: Find local minimum}
\State ${\bf w}^* \gets$ \Call{Vanilla-Grad-Desc}{${\bf w}$}

\Statex \hspace{1cm}$\rhd$\textmd{ Stage 2: Sparsify {\bf w} }
\hspace{2cm}
\State   Assign initial near zero weights $\{w_i\}$ to Laplacian cluster, $C_l$, or to Gaussian cluster, $C_g$ using posterior.
\State Estimate priors, $\{p(C_l({\bf w})),p(C_g({\bf w})\}$
\Repeat
\hspace{1cm}
\State  $\{\pi_g, \pi_l,\sigma_g,b_l\} \gets $ hyper-parameter updates
\State ${\bf w}^* \gets$ model updates using (\ref{eqn:gauss_l_mixtureX}) 
\State  Estimate posteriors $\{p(C_l|{\bf w}^*)$,$p(C_g|{\bf w}^*)\}$
\State Update clusters, $C_l$ and $C_g$ 
%
\Until{No clusters change, or a finite \# of iterations completed}

\Statex \hspace{1cm}$\rhd$\textmd{ Stage 3: Probe each dimension in a radius R}
\Statex \hspace{1cm}$\rhd$\textmd{ to escape local minimum (if possible)}
%
%
\For{$i=1\ldots d$}\Comment{For each dimension, $w_i$}
\For{$step \in \{\epsilon_S,-\epsilon_S,2\epsilon_S,-2\epsilon_S,\dots,R,-R\}$}
\State  ${\bf w} \gets {\bf w}^*$
\State $w_i \gets w_i + step$
\If{$L_{\gamma}({\bf w}^*)-L_{\gamma}({\bf w}) \geq \epsilon_L$}
\State \textbf{break} \Comment{Goto step 3}
\EndIf
\EndFor 
\EndFor
\Until{$L_{\gamma}({\bf w}^*)-L_{\gamma}({\bf w}) < \epsilon_L$}
\State \textbf{return} ${\bf w}^*$
\EndFunction
\end{algorithmic}

\end{algorithm}

\end{document}